\documentclass{article}

\usepackage[preprint]{neurips_2021}

\PassOptionsToPackage{authoryear}{natbib}
\usepackage{xcolor}         
\usepackage[utf8]{inputenc} 
\usepackage[T1]{fontenc}    
\usepackage{hyperref}       
\usepackage{url}            
\usepackage{booktabs}       
\usepackage{amsfonts}       
\usepackage{nicefrac}       
\usepackage{microtype}      
\usepackage{lipsum}
\usepackage{graphicx}
\usepackage{amsmath}
\usepackage{amssymb}
\usepackage{multirow}
\usepackage{makecell}
\usepackage{bm}
\usepackage{comment}

\graphicspath{ {./images/} }

\newcommand{\w}{\boldsymbol{\theta}}
\newcommand{\m}{\boldsymbol{m}}

\title{Towards Understanding Iterative Magnitude Pruning: Why Lottery Tickets Win}

\author{
Jaron Maene \\
  KU Leuven\\
  \texttt{jaron.maene@student.kuleuven.be} \\
   \And
Mingxiao Li \\
  KU Leuven\\
  \texttt{mingxiao.li@kuleuven.be} \\
  \And
Marie-Francine Moens \\
  KU Leuven\\
  \texttt{sien.moens@kuleuven.be} \\
}

\begin{document}

\maketitle

\begin{abstract}
  The lottery ticket hypothesis states that sparse subnetworks exist in randomly initialized dense networks that can be trained to the same accuracy as the dense network they reside in. However, the subsequent work has failed to replicate this on large-scale models and required rewinding to an early stable state instead of initialization. We show that by using a training method that is stable with respect to linear mode connectivity, large networks can also be entirely rewound to initialization. Our subsequent experiments on common vision tasks give strong credence to the hypothesis in \cite{evci_gradient_2020} that lottery tickets simply retrain to the same regions (although not necessarily to the same basin). These results imply that existing lottery tickets could not have been found without the preceding dense training by iterative magnitude pruning, raising doubts about the use of the lottery ticket hypothesis.\footnote{Our code is available at \url{https://github.com/jjcmoon/understanding_imp}.
}
\end{abstract}

\section{Introduction}
Pruning is an established method for significantly reducing the parameter count of a dense neural network \citep{gale_state_2019, blalock_what_2020}. The basic template for pruning consists of first training a dense network, then
removing weights according to some heuristic, such as magnitude. For the best results, this can be iterated, alternatingly pruning weights and retraining the network \citep{renda_comparing_2019}. 

Pruning still requires training the full network first; pruning offers a way towards more efficient inference but not more efficient training. \cite{frankle_lottery_2019} suggest that sparse networks could be trained from scratch by introducing the lottery ticket hypothesis (LTH). This hypothesis states that every dense network contains a subnet that can be trained to equal (or higher) accuracy than the full network in equal (or fewer) training epochs. These successful sparse subnets are called lottery tickets. \cite{frankle_lottery_2019} retroactively find lottery tickets using iterative magnitude pruning (IMP). This works similarly to regular pruning, with the difference that after the weights are removed, the remaining weights are reset to their initial values before retraining. For more details on IMP, we refer to the extensive discussion in \citet{frankle_lottery_2019}.

\textbf{Stability.} Subsequent work \citep{liu_rethinking_2018, gale_state_2019} could not find lottery tickets for large vision models. \citet{frankle_linear_2020} resolve this by slightly weakening the hypothesis: networks cannot necessarily be reset to initialization but instead to some early phase in training. The authors show that this corresponds with network stability: resetting works only when resetting to a stable network state. 

The trajectory of stochastic gradient descent (SGD) is inherently noisy because of random data ordering and, for example, because of randomized data augmentations. \citet{frankle_linear_2020} quantify this stability using linear mode connectivity. A network is trained twice with different SGD noise, after which the network is called stable if the two solutions are connected with a linear path of low loss.

We propose an alternative method of stabilizing the LTH. Instead of rewinding to an early epoch, we simply stabilize the training procedure itself. Although this could be achieved in several ways, we focus on increasing the batch size. In this way, we can find lottery tickets for large networks from initialization. In line with \citet{frankle_linear_2020}, stabilizing training induces linear mode connectivity, at the same time as when lottery tickets start to manifest.

\textbf{IMP Interpretation.} \citet{evci_gradient_2020} recently proposed a possible interpretation for the behavior of IMP and the success of the LTH. The authors posit that lottery tickets cannot be considered random initializations but that a lottery ticket contains a prior for rediscovering the solution of the model from which it was pruned. We call this hypothesis the \textit{regurgitating tickets interpretation} (RTI): a lottery ticket retrains to a similar optimum compared to the network from which it was pruned. As discussed in \cite{evci_gradient_2020}, the RTI allows us to informally understand several key questions regarding IMP, such as why lottery tickets transfer or why they are robust to perturbations. 

To empirically validate the RTI, \citet{evci_gradient_2020} evaluate the similarity of trained lottery tickets on LeNet and ResNet50. Although enticing, the authors perform the evaluation only at a single sparsity level. Moreover, this does not conclusively prove the RTI, as alternative explanations remain possible. Solutions from successive pruning rounds might merely be close as a side effect of the very stable training environment or other properties of the lottery tickets.

We design a new experiment to affirm the causal relationship between the success of IMP and the RTI across sparsities and on various vision architectures. We modify the loss function to repel previously found optima. We show that this removes the ability of IMP to find lottery tickets. On the other hand, this has no noticeable effect on the performance of random reinitalizations.

Finally, we explore what notion of similarity is entailed by the RTI and how it is affected by the stability of training. We show that when IMP becomes stable, all pruning rounds have a very small angular distance. \cite{evci_gradient_2020} make the stronger claim that IMP retrains into the same basin. We find that this is not necessarily the case at very low or high sparsities.

The LTH has spawned the idea that lottery tickets could be generated in a mostly data-independent fashion without relying on the very costly IMP. This would open up the possibility of fully sparse training without any prior dense training. For example, \cite{morcos_one_2019} hint at \textit{"the possibility of parameterizing the distribution of such tickets, allowing us to sample generic, dataset-independent winning tickets."} Our evidence for the RTI indicates that such efforts might be misguided, as the construction of a lottery ticket requires prior knowledge of a dense solution. This is in line with \cite{frankle_pruning_2020}, who empirically examine current attempts at data-independent pruning at initialization (PAI), and show that these attempts significantly underperform compared to IMP. 

We hope that an improved understanding of IMP and the LTH, will allow future work to design better pruning and sparse optimization algorithms. In summary, our key contributions are as follows:

\begin{enumerate}
 \item We demonstrate that the lottery ticket hypothesis can be extended to large networks and show that \textit{it is possible to train large neural networks from scratch without rewinding} by using a sufficiently large batch size. This reinforces the hypothesis of \citet{frankle_linear_2020} that finding lottery tickets is possible when training is stable with respect to linear mode connectivity.

 \item We introduce the \textit{regurgitating tickets interpretation} (RTI) based on the prior work in \cite{evci_gradient_2020} and test its causal relationship with the success of IMP using repellent and attractive losses. Using the RTI, we can explain the relation of IMP with stability. Furthermore, we can make testable predictions regarding the behavior of IMP, such as the effect of weight decay in section~\ref{sec:weight_decay}.

 \item We characterize the similarity of the pruned and retrained networks with the original networks. We show that similarity is strongest on the medium-sparsity high-accuracy plateau. Very low and high sparsities are not connected by a linear low error path, in contrast to \cite{evci_gradient_2020}, although the angular distances are still considerably smaller than those of reinitializations.
\end{enumerate}

\section{Alternative stabilization of the lottery ticket hypothesis}\label{sec:big_batch}

\cite{frankle_linear_2020} link the applicability of IMP with the stability of training. The authors use this connection to stabilize the LTH by showing that during training, there exists a period after which the network becomes stable. We consider the complementary property that every training method can be made stable from initialization by using a sufficiently large batch size. 

This claim should certainly not be surprising: in the limit, increasing the batch size results in nonstochastic full-batch gradient descent, which by definition does not exhibit gradient noise. Hence, sufficiently large batch training is always stable (disregarding data augmentations). However, this allows us to view the success of IMP as a property of the training procedure and not as a property of the network. In other words, we hypothesize that lottery tickets do exist in all networks; a failure to find them is not proof of their absence but rather an indication that IMP is unstable. In later sections, we discuss how this relates to the RTI. 

\subsection{Experiments}

There are several possibilities for reducing the instability during training, with decreasing the learning rate and increasing the batch size being the most straightforward. We mostly focus on the latter, as the added data parallelism is useful in practice. Another method of interest is gradual warm-up: the linear increase in learning rate at the start of training, originally introduced for large-batch training in \citet{goyal_accurate_2018}.
This method specifically targets the instability in early training, which has been shown to be the most severe \citep{frankle_early_2020}. \citet{frankle_lottery_2019} has already shown that warm-up combined with a low learning rate can stabilize IMP on some networks like VGG19 and ResNet20.

We perform IMP by iteratively training and pruning weights for 25 rounds. Before starting a new round, the remaining weights are reset to the original initialization. The hyperparameters used for training are summarized in table~\ref{table:hyperparams}. More details on the training setup can be found in appendix~\ref{sec:training_setup}. 

\begin{figure}
\centering
\includegraphics[width=\textwidth]{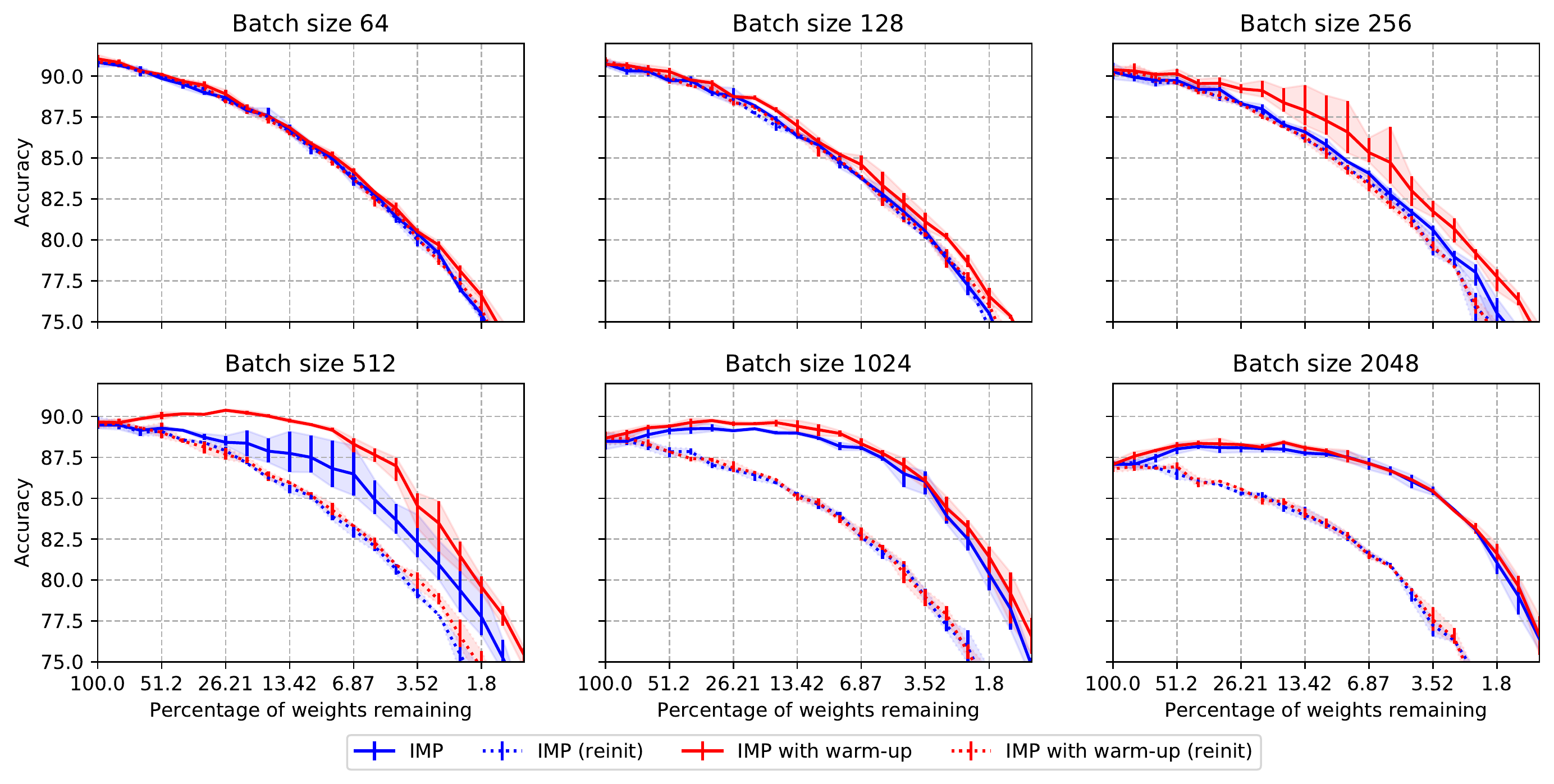}
\caption{Influence of the batch size and warm-up on IMP. Experiments on CIFAR-10 with ResNet20. Average of 3 runs with different seeds; error bars indicate maximum and minimum values.}\label{figure:big_batch_sweep}
\end{figure}

Figure~\ref{figure:big_batch_sweep} displays the effect of varying the batch size and adding a long warm-up (10 epochs). Using smaller batch sizes, IMP is not superior to random reinitializations, while it is with larger batch sizes. The increase does not have a binary effect on IMP. There is a transitional semistable region around batch size 512. The additional stability of the warm-up helps during this transition, where the training is barely stable enough. In a high batch size training regime it is already fully stable; hence, adding the warm-up has little effect. We also discuss this stabilization on Conv-4 and ResNet50 with similar results; see appendix~\ref{app:quantifying similarity}.

\begin{figure}
\centering
\includegraphics[width=\textwidth]{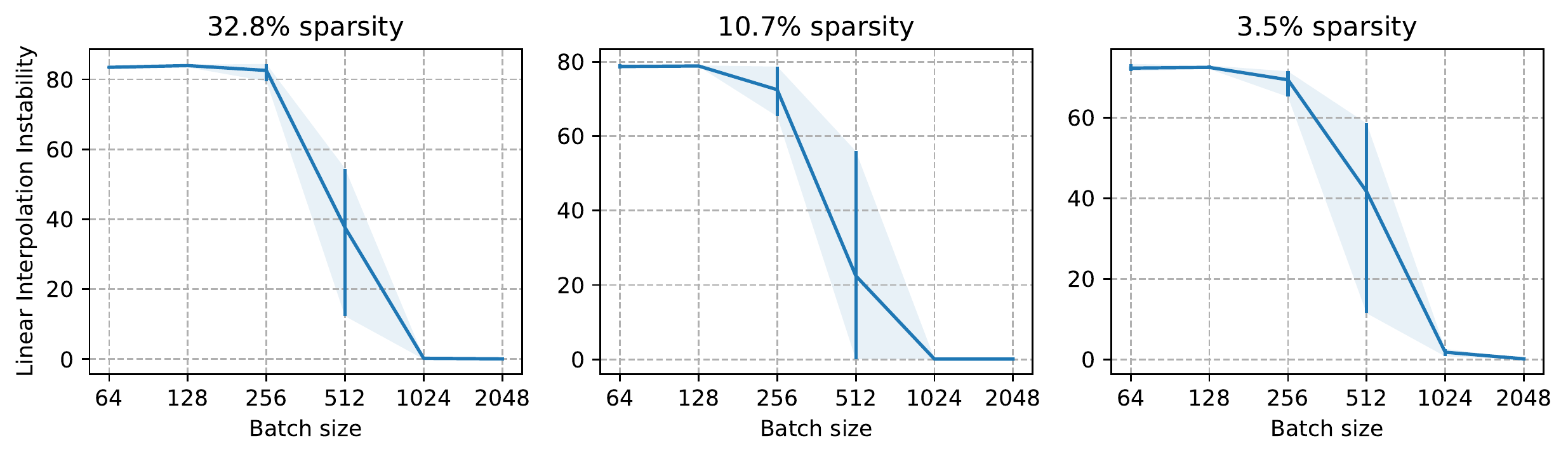}
\caption{Influence of the batch size on the linear interpolation instability at pruning rounds 5, 10 and 15. Experiments on CIFAR-10 with ResNet20, without warm-up. Average of 3 runs with different seeds; error bars indicate maximum and minimum values.}\label{figure:LII}
\end{figure}

We find that, as in \cite{frankle_linear_2020}, IMP can find winning tickets only when the subnetworks have low linear mode instability (see figure~\ref{figure:LII}). To test this, we train each network twice with different SGD noise (i.e., data ordering and augmentations). The linear mode instability is the maximal error increase on the linear path between these two solutions. For a more rigorous introduction to linear mode connectivity and instability,  refer to \citet{frankle_linear_2020}.

\begin{table*}
\scriptsize
\centering
\begin{tabular}{l@{\ }l@{\ }|@{\ }c@{\ }c@{\ } c@{\ }c@{\ \ }c@{\ \ }c@{\ \ }c@{\ }c@{\ } c@{\ } c@{\ } @{\ }c@{\ }c@{\ }}
\toprule
Network & Dataset & Params& Epochs & Batch & Optimizer & LR & Schedule & Weight Init & WD & Pruning Style \\ \midrule
LeNet & MNIST & 266K & 25 & 60 & Adam &  12e-4 & constant & Kaiming uniform & 0 &  Iterative \\ \midrule
ResNet20 & CIFAR-10 & 274K & 85 & 128 & momentum & 0.1 & 10x drops at 56,71 & Kaiming normal & $10^{-5}$ &  Iterative \\ \midrule
Conv-4 & CIFAR-10 & 2.4M & 75 & 128 & momentum & 0.01 & cosine & Kaiming normal & $10^{-5}$ &  Iterative \\ \midrule
ResNet50 & ImageNet & 25.5M & 90 & 1024 & momentum & 0.4 & 10x drop at 30,60,80 & Kaiming normal & $10^{-5}$ &   One-shot (30\%) \\
\bottomrule
\end{tabular}
\caption{Networks and hyperparameters, based on \cite{frankle_linear_2020}. LR denotes learning rate, and WD is weight decay. For more information on the choices of datasets and architectures, see appendix~\ref{sec:training_setup}.}
\label{table:hyperparams}
\vspace{-1mm}
\end{table*}

\section{Forced repellence of previous minima}\label{sec:poison}

To test the causal relationship between the RTI and the success of IMP, we design an experiment where IMP is forced to train to a different optimum at each pruning round. We accomplish this by modifying the loss function: an extra term $J_r(\w)$ that repels the current weights $\w$ from any previous optima is added to the loss. We denote the mask of the current sparse network $\m$. Given the set of previous optima $\Theta$, we define the repellence loss as the average of the squared cosine similarities to these minima:

\[ J_r(\w) = \frac {\lambda_r} {|\Theta|} \sum_{\w_r \in \Theta} \left( \frac {(\w \odot \m) \cdot (\w_r \odot \m)} {|| \w \odot \m ||_2 \cdot || \w_r \odot \m||_2} \right)^2 \]

The hyperparameter $\lambda_r$ controls the strength of the repellence. We note that in practice, the specific value of $\lambda_r$ is not very important. As long as $\lambda_r$ is sufficiently large, the network learns an orthogonal solution, and the repellent loss becomes negligible after training. We employ a default value of $\lambda_r=2$; this parameter is tuned on CIFAR-10 to be just sufficient to repel the model from all previous optima.

\begin{figure}
\centering
\includegraphics[width=\textwidth]{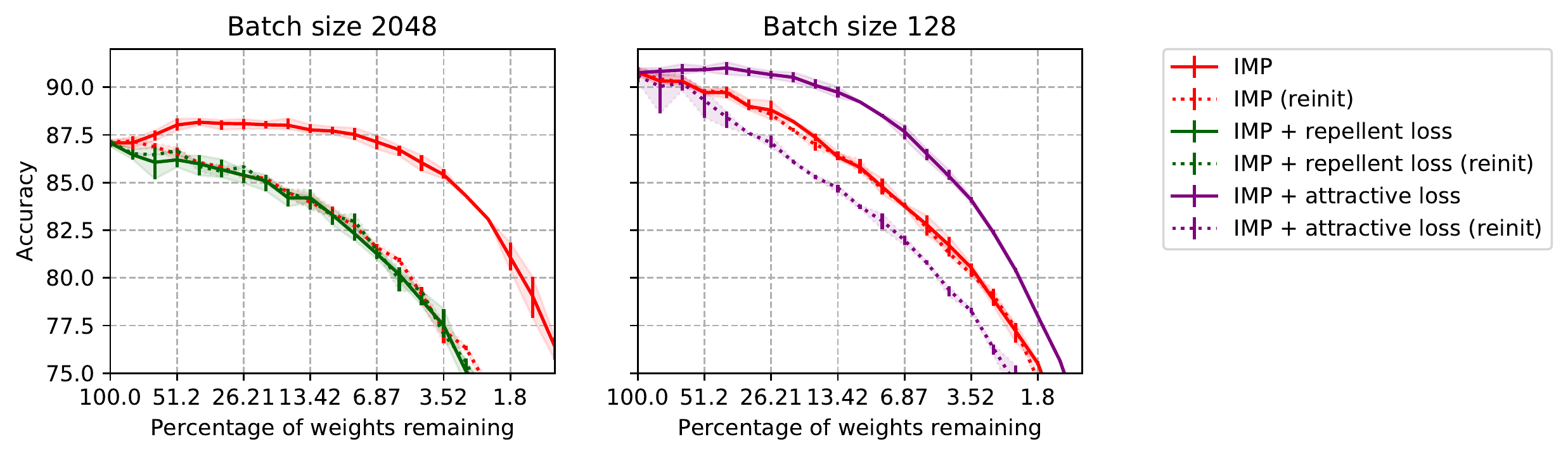}
\caption{Comparison of repellent IMP with regular IMP. (left): stable IMP with repellent loss ($\lambda_r=2$). (right): unstable IMP with attractive loss ($\lambda_r=-1$). Experiments on CIFAR-10 with ResNet20, without warm-up. Average of 3 runs with different seeds; error bars indicate maximum and minimum values.}\label{figure:IMP_repellent}
\end{figure}

\subsection{Experiments}
We repeat a stable IMP experiment from section~\ref{sec:big_batch} with the repellent loss; the results are visualized on the left of figure~\ref{figure:IMP_repellent}. The addition of the repellent loss removes the ability to find lottery tickets, and induces performance similar to that of the random reinitialization of regular IMP. In contrast, the reinitalization performance is not noticeably affected. Experiments on LeNet and Conv-4 can be found in appendix~\ref{app:forced_repellence}, but exhibit the same behavior.

We can also perform the reverse exercise: by making $\lambda_r$ negative, we attract the network to previous solutions. This allows previously unstable training to retrain to these previous solutions; see the right of figure~\ref{figure:IMP_repellent}. Although the attractive experiment can find lottery tickets, both IMP and its reinitialization do not fully perform on par with, for instance, rewinding. This is expected: training as close as possible to the previous solution within the limits of the subnet is too strong of a constraint. This is even detrimental in the case of re-initializations: when the previous solution region lies far from the mask subspace, the attraction is counterproductive.

A downside of our formulation of the repellence loss is that it incurs memory overhead at higher sparsities due to the growth of $\Theta$ with every pruning round. However, a simpler design that repels only from the previous optimum is not sufficient. IMP circumvents this in a biphasic manner: the network just retrains to the last but one optima. The resulting checkerboard similarity pattern can be found in figure~\ref{figure:heatmap_checkerboard} in appendix~\ref{app:forced_repellence}.

\section{Quantifying the similarity between IMP rounds}\label{sec:similarity}

In the formulation of the RTI, we informally characterize the successive optima from IMP as similar. In this section, we attempt to paint a more in-depth quantitative picture of what this similarity entails and its relation with sparsity and stability.

\subsection{Angular distances}\label{sec:angular_distances}

To compare networks at different pruning rounds, we define the distance between two networks of different sparsities as the angle between their shared weights. More formally, take two sparse networks with weight vectors $\w_1$ and $\w_2$ and with masks $\m_1$ and $\m_2$, respectively. We define the distance between these networks as $d(\w_1 \odot \m, \w_2 \odot \m)$, where $\m = \m_1 \odot \m_2$ is the intersection of the connections in the two networks. Note that in our case, these networks are the result of iterative pruning, and hence, $\m_2 = \m$ is always a strict subset of $\m_1$.

\begin{figure}
\centering
\includegraphics[width=\textwidth]{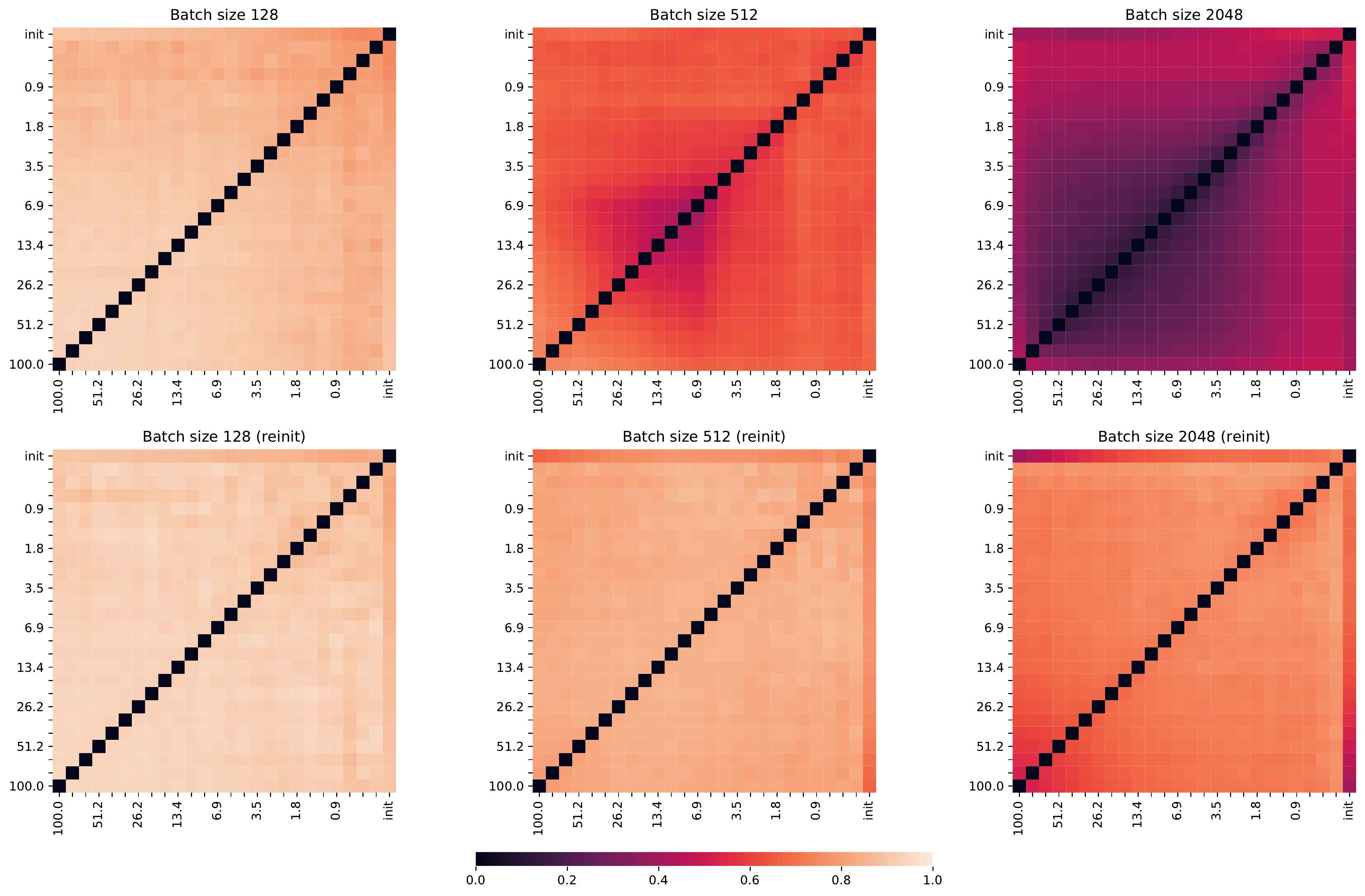}
\caption{Angular distances between the retrained networks throughout sparsity. The last row (and column) is a comparison with the network at initialization. Average of 3 runs with different seeds. Experiments on CIFAR-10 with ResNet20, without warm-up.}\label{figure:heatmaps}
\end{figure}

For our purposes, we take the angle between networks as the distance function: $d(\mathbf{v}_1, \mathbf{v}_2) = 2 \cdot \angle(\mathbf{v}_1, \mathbf{v}_2) / \pi$. An angle close to 0 means that the networks are very similar. Unrelated networks are almost orthogonal due to the large dimensionality and hence have an angular distance close to 1. Other choices for distance functions such as cosine distance or Euclidean distance are explored in appendix~\ref{sec:distance_functions} but deliver similar results. An advantage of angular distances is that they are invariant to the model norm, which is typically larger for a sparse network than a subset of a dense network.

Figure~\ref{figure:heatmaps} displays the distances between the pruning rounds by the IMP experiments from figure~\ref{figure:big_batch_sweep}. At smaller batch sizes, such as the conventional 128, the networks retrain to almost orthogonal networks every iteration, as does it with the reinitialization. However, at large batch sizes (here 1024 and 2048), the networks are separated by only a very small angle. We observe a clustering of especially small angles at medium sparsities. As noted in \cite{frankle_linear_2020}, the medium sparsities stabilize first. These regions align with the high-accuracy plateau of IMP at medium sparsities (see figure~\ref{figure:big_batch_sweep}). Appendix~\ref{app:quantifying similarity} contains the same visualizations for the other architectures.

The reinitializations remain fairly orthogonal with different batch sizes. However, at large batch sizes, there is a slight homogeneous decrease in the distance. This can be explained by the fact that they have been trained for fewer steps because the batch size is increased without increasing the training epochs. Hence, all solutions stay somewhat close to the initialization (compare the last row/column of each matrix in figure~\ref{figure:heatmaps}).

\subsection{Pruning thresholds and weight decay}\label{sec:weight_decay}

\begin{figure}
\centering
\includegraphics[width=\textwidth]{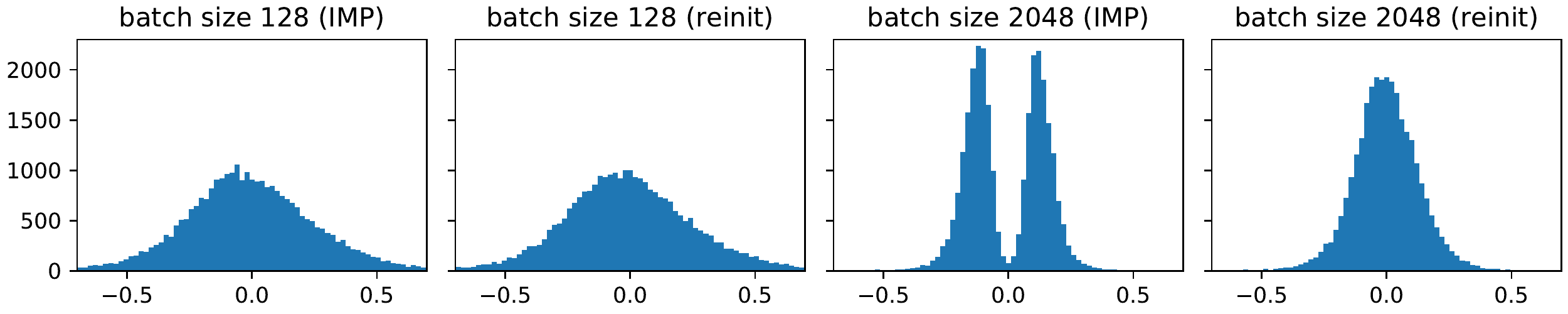}
\caption{Histograms of the model weights after 10 pruning iterations (at sparsity 10.7\%). Experiments on CIFAR-10 with ResNet20, without warm-up.}\label{figure:histograms}
\end{figure}

A corollary of the RTI is that the average magnitude of the weights start increasing through pruning rounds. Indeed, small weights are pruned, and the remaining weights retrain to similar values. See figure~\ref{figure:histograms} for a visualization of the weight histograms of IMP. In the stable case, the network trains to an optimum with few weights around zero, giving rise to a bimodal distribution. Note that this occurs even though weight decay is active during training.

\begin{figure}
\centering
\includegraphics[width=\textwidth]{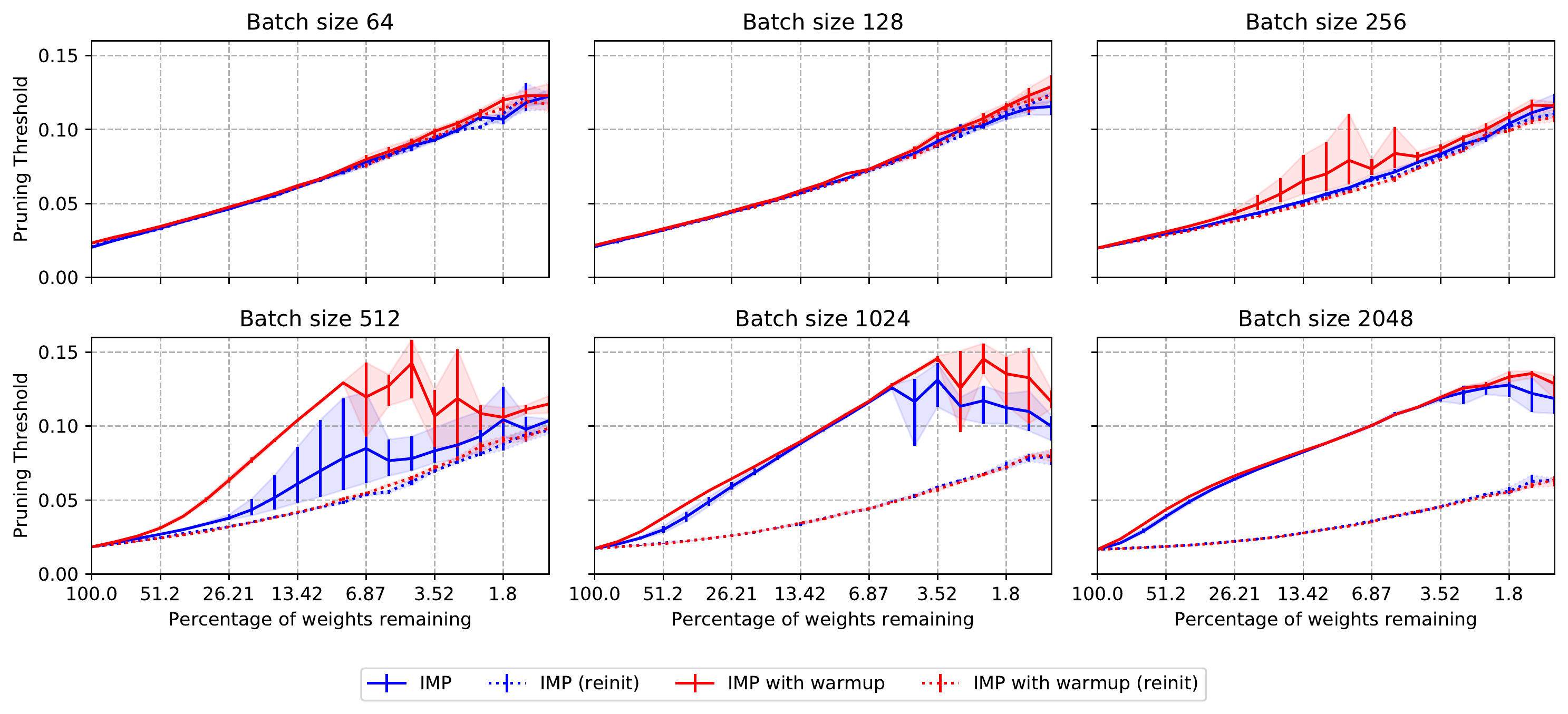}
\caption{Influence of the batch size, warm-up and sparsity on the pruning threshold using IMP. Experiments on CIFAR-10 with ResNet20. Average of 3 runs with different seeds; error bars indicate maximum and minimum values.}\label{figure:thresholds}
\end{figure}

A further consequence of this increase in the weight magnitude is that IMP has a higher pruning threshold than reinitializations (see figure~\ref{figure:thresholds}). Observe how stable IMP induces an above random increase in the pruning threshold, which later collapses again for semistable settings.

\begin{figure}
\centering
\includegraphics[width=.9\textwidth]{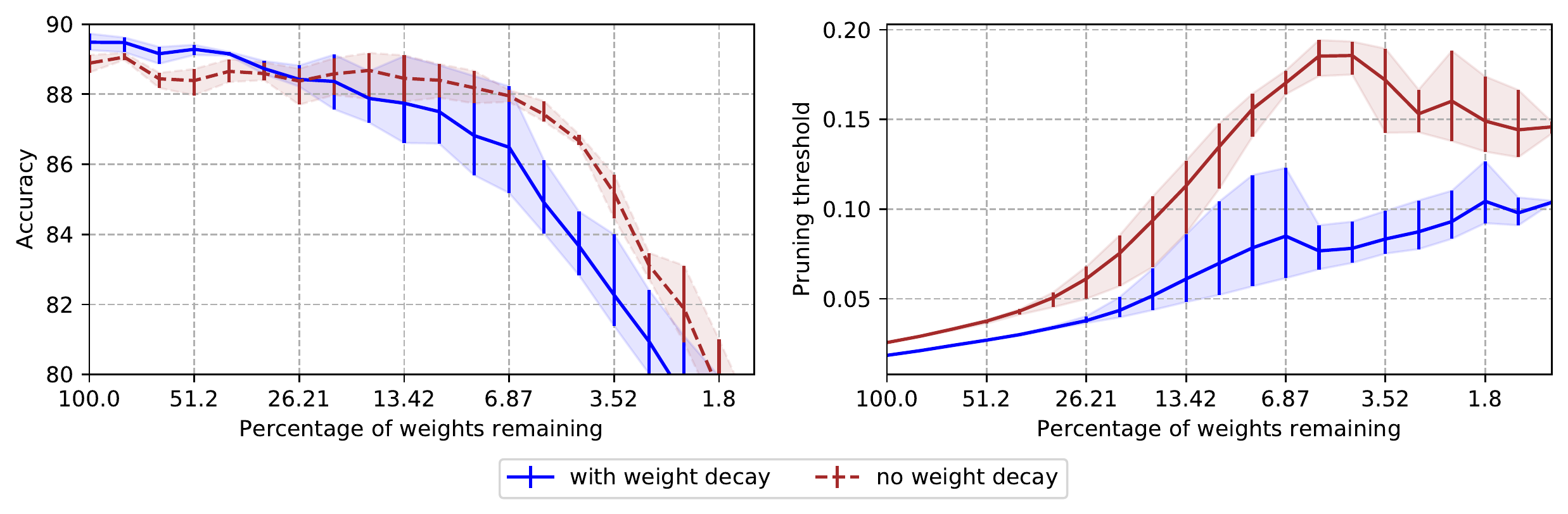}
\caption{Influence of weight decay in semistable IMP on the accuracy (left) and the pruning threshold (right). Experiments on CIFAR-10 with ResNet20, with a batch size of 512 and no warm-up. Average of 3 runs with different seeds; error bars indicate maximum and minimum values.}\label{figure:no_weight_decay}
\end{figure}

The above discussion implies that weight decay might be disadvantageous for IMP. Indeed, it can push the solution away from its previous (high magnitude) values. We confirm this in figure~\ref{figure:no_weight_decay}, where we repeat semistable IMP runs without weight decay. At low sparsity, this slightly reduces performance due to the lack of regularization, but from there, the pruning threshold grows faster and collapses later, resulting in a longer high-accuracy plateau. In the fully stable setting, this is less of a concern, as the threshold does not collapse (see figure~\ref{figure:thresholds} bottom right).

\subsection{Error barriers}\label{sec:error_barrier}

\cite{evci_gradient_2020} claim that lottery tickets lie in the same basin as the solution they are pruned from. They base this on the linear interpolations between optima. However, their analysis is limited to considering only a single sparsity.

\begin{figure}
\centering
\includegraphics[width=0.9\textwidth]{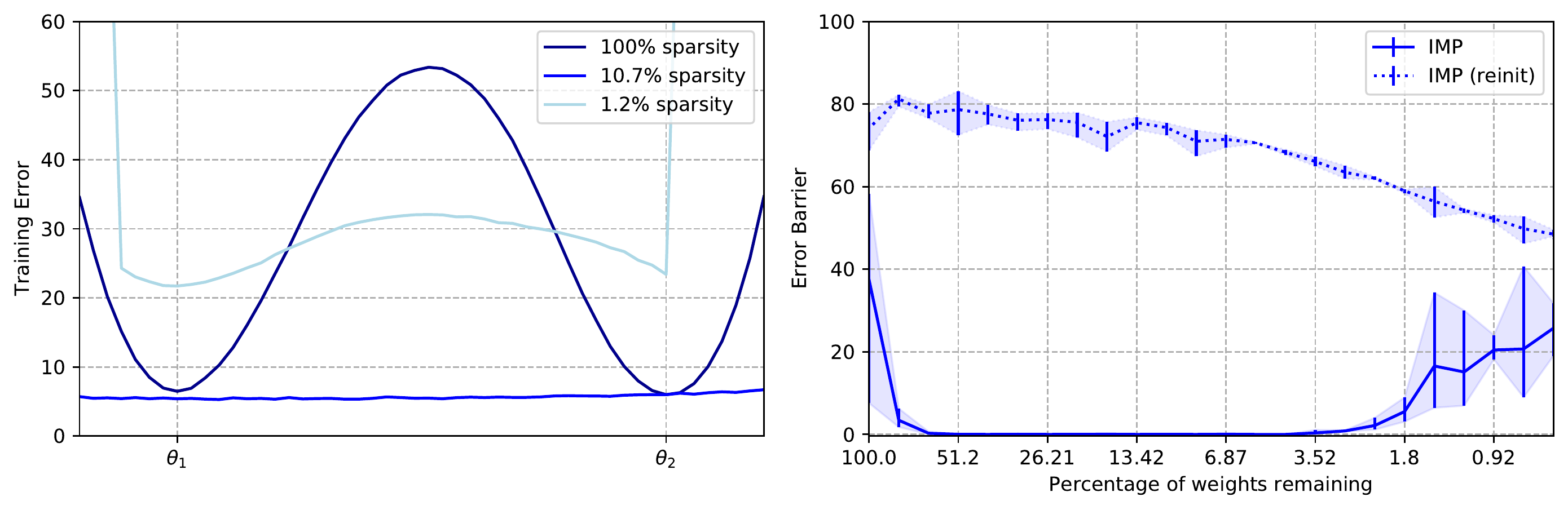}
\caption{(left): Linear interpolations between two successive pruning iterations for three different sparsities. (right): The maximum error increase on the linear path over sparsities. Experiments on CIFAR-10 with ResNet20, with a batch size of 2048 and no warm-up. Average of 3 runs with different seeds; error bars indicate maximum and minimum values.}\label{figure:error_barrier}
\end{figure}

Their study of linear paths is founded on the work of \cite{frankle_linear_2020}, which extensively discusses linear paths in connection with the stability of IMP. We use the same approach to look at the linear paths between the optima of two successive pruning rounds. On the left side of figure~\ref{figure:error_barrier} we visualize the linear interpolation between low-, medium- and high-sparsity networks (at $\theta_1$) with a pruned and retrained network (at $\theta_2$). The right of figure~\ref{figure:error_barrier} shows the size of the error barrier (the maximal increase on the linear path) for every sparsity. To calculate this, the linear path is equidistantly sampled 20 times. At medium sparsity, the modes are indeed linearly connected, while in the very low and high sparsity settings, this is not the case. These observations correspond with figure~\ref{figure:heatmaps}, which also shows that optima are most similar at medium sparsities. This experiment is repeated on LeNet and Conv-4 in appendix~\ref{app:quantifying similarity}.

\section{Related work}\label{sec:related_work}

\textbf{Lottery ticket hypothesis}. \cite{frankle_lottery_2019} demonstrate the existence of lottery tickets: sparse neural networks hidden inside a dense randomly initialized network that can be trained to comparable or higher accuracy than the dense model. \citet{liu_rethinking_2018} and \citet{gale_state_2019} question the LTH by showing that these lottery tickets cannot be found in large-scale models such as ResNet50. \cite{frankle_linear_2020} resolve this by adjusting the LTH with rewinding. The sparse networks (called matching tickets instead of lottery tickets) cannot be completely reset to initialization but can be set to an early state in training. They show that this corresponds with the stability of the training. As soon as the network becomes stable to the noise of SGD, it is safe to reset to. \cite{zhou_deconstructing_2019} further investigate the LTH using ablations and show that lottery tickets can already achieve nontrivial sparsities without training. \cite{evci_gradient_2020} explain that the success of lottery tickets lies in relearning the same solution as the larger net that they were pruned from.

\textbf{Large-batch training}. There has been significant recent interest in large-batch training, as it allows massive data parallelism. \citet{goyal_accurate_2018} show that large vision models on ImageNet can be trained using large batches without a decrease in accuracy, by using gradual warm-up and linearly increasing the learning rate with the batch size. \citet{mccandlish_empirical_2018} relate training stability with the efficacy of large-batch training. They show that increasing the batch size stabilizes the gradient up to a task-dependent limit: the gradient noise scale. After this, the returns diminish. 

\textbf{Pruning at initialization}. Recent works have proposed pruning a network all the way at initialization, avoiding any dense training. Examples include SNIP \citep{lee_snip_2018}, GraSP \citep{evci_gradient_2020} and SynFlow \citep{tanaka_pruning_2020}. \cite{frankle_pruning_2020} question the effectiveness of these methods. They showed that randomly shuffling the connections in the mask produced by the PAI methods does not decrease the performance. This means that the per-layer density of the weights is sufficient to explain the performance of the current PAI methods.


\section{Discussion}\label{sec:discussion}

\textbf{Regurgitating tickets interpretation.} Using experiments with repellent and attractive losses, we are able to selectively enable and disable the finding of lottery tickets. 
Hence, lottery tickets found with IMP cannot be seen as evidence of the feasibility of training sparse networks from scratch without pretraining. This raises doubts about the feasibility of data-independent pruning or PAI methods.

Of course, there are still many other promising avenues that could lead to efficient sparse training, for example, transferring existing lottery tickets \cite{morcos_one_2019, mehta_sparse_2019}, pruning weights during training \citep{you_drawing_2019}, or dynamically changing the mask during training \citep{evci_rigging_2020, savarese_winning_2020, dettmers_sparse_2019,kusupati_soft_2020}.


\textbf{Stability.} The RTI implies an obvious explanation for the need for IMP for stability: only when the network is stable enough to retrain to a similar region can we find lottery tickets. Our results in section~\ref{sec:big_batch} confirm this by looking at the large-batch training on common vision tasks. This reinforces the hypothesis of \citet{frankle_linear_2020} that finding lottery tickets is possible only when resetting to a state that is stable to SGD noise.

The main advantage of stabilizing training using rewinding \citep{frankle_linear_2020} is that it does not require changing any hyperparameters. Our large-batch experiments modified only the batch size and correspondingly are not competitive with rewinding. Although this might be fixable by an expensive hyperparameter search or longer training, it is unclear whether this would make sense. Given the RTI, there are no clear advantages to using IMP with resetting, instead of ordinary iterative pruning.

Here we only considered stabilizing SGD with momentum. Adaptive optimizers such as Adam could form a further stable alternative to rewinding or large-batch training, as they typically employ much lower learning rates. Though a full discussion is beyond the scope, \cite{chen_lottery_2020} evaluate the LTH on an Adam-optimized BERT and find winning tickets without rewinding.

\textbf{Similarity of RTI.} The loss landscapes of neural networks are a complex subject, and characterizing the similarity of the pruned networks to their previous solutions requires nuance. We find that they do not necessarily lie in the same basin at very low or high sparsities. Also they are not simply the solution that is as close as possible to the previous optimum. On the other hand, the angles between all rounds in the IMP are very small compared to re-initializations, indicating the solutions are still related.
We hope to lay a more rigorous foundation for this in future work.

\textbf{Evidence-based pruning.} As a more concrete application of our work, our empirical methods to analyze the behavior of IMP might be useful for improving current pruning techniques.

For example, \cite{renda_comparing_2019} show that complete retraining is superior to just fine-tuning when pruning iteratively. Our results on mode connectivity suggest that a compromise might be possible. Fully retraining on the first couple of rounds with the linear mode connectivity at medium sparsities could permit less expensive training schedules during the rest of pruning.
Another example is that weight decay might not be desirable for pruned models. For one, regularization techniques make less sense in undercapacitated models. But also because the weight decay pushes the solution away from the good solution found in dense training.

\section{Limitations and broader impact}\label{sec:limitations}
In line with previous works, we studied only supervised computer vision tasks with dense and convolutional architectures. Due to practical constraints, we refrained from experimenting on very large models such as vision transformers.  
Our analysis of the RTI considered only lottery tickets generated by IMP. However, it is not impossible that other methods to find lottery tickets exist which do not suffer from the RTI, and our discussion of RTI should hence not necessarily be taken as an absolute statement on the LTH but only of the LTH in relation to IMP.

Because of the rather foundational nature of our work, the direct impact of our work is limited. We hope that an improved understanding of lottery tickets might lead to more efficient sparse models. We note that caution should be taken when compressing models for sensitive domains, as it can impact the bias and fairness of the model. For example, \citet{hooker_what_2020} show that a reduction in capacity of the model can severely affect the accuracy of a small subset of classes. 

\section{Conclusion}

We showed that sparse large vision networks can be trained from scratch by using larger batch sizes. As soon as this stabilizes training enough to achieve negligible linear interpolation instability, we can find lottery tickets. 
Based upon \cite{frankle_linear_2020} and \cite{evci_gradient_2020}, we hypothesized and tested the \textit{regurgitating tickets interpretation} (RTI) as an explanation for the lottery ticket hypothesis. It states that IMP works by retraining the same initializations to similar optima throughout pruning rounds. 
We showed that there are quantitative differences in this similarity between low/high sparsities and medium sparsities. Medium sparsities stabilize first and are linearly connected to the model from which they are pruned.
Finally, we discussed the possible implications of the RTI for applications of the lottery ticket hypothesis, such as the infeasibility of data-independent pruning approaches.

\begin{ack}
The authors have no competing interests. The authors did not receive funding for this research.

\end{ack}

\bibliographystyle{apalike}
\bibliography{neurips_2021} 

\newpage
\appendix

\section*{Appendix overview}

In appendix~\ref{sec:training_setup} we discuss some additional technical details concerning the experiments. Appendix~\ref{sec:distance_functions} compares the choices of distance functions to quantify the similarity between sparse networks (see section~\ref{sec:angular_distances}). In appendix~\ref{app:alternative_stabilization} we display the additional experiments on the alternative stabilization from section~\ref{sec:big_batch}. Appendix~\ref{app:forced_repellence} does the same for the forced repellence experiments from section~\ref{sec:poison}, and appendix~\ref{app:quantifying similarity} for the similarity quantification from section~\ref{sec:similarity}. 

\section{Experiment setup}\label{sec:training_setup}

\subsection{Datasets and architectures}

We experimented on the following 3 standard vision datasets:
\begin{itemize}
  \item MNIST \citep{lecun_gradient-based_1998}. Available under the Creative Commons Attribution-Share Alike 3.0.
  \item CIFAR-10 \citep{krizhevsky_learning_2009}. Available under the MIT license.
  \item ImageNet (Face-blurred ILSVRC 2017) \citep{deng_imagenet_2009}. Available for research purposes, not available under a single license.
\end{itemize}

We chose these datasets due to their prevalence in previous literature on pruning and the lottery ticket hypothesis, making comparison easier. The above license information was determined on best effort by the authors.

In terms of the architectures, we present results on 4 different vision models of varying scales. The smallest model is LeNet. This is a very basic dense network, with 3 layers (width 300 - 100 - 10). Conv-4 is a small VGG-like convolutional network, taken from \cite{frankle_lottery_2019}. It has 4 convolutional layers interspersed by 2 max-pooling layers, followed by 3 dense layers. The ResNet family \citep{he_deep_2016} has convolution layers with residual connections and batch normalization, enabling much deeper networks.

\subsection{Hardware}

All training was performed on our own hardware, namely two GPUs (NVIDIA GeForce GTX TITAN X). On this GPU a single ResNet20 experiment on CIFAR-10 takes about half an hour (depends on hyperparameters like batch size). For each IMP experiment, we train the network 25 times. So replicating e.g. figure~\ref{figure:big_batch_sweep} requires training the network 1800 times, costing in the order of 37 GPU days.

\subsection{Pruning}

All pruning is global over all layers. Only weights are pruned explicitly. Biases get pruned implicitly; if a neuron has no connections anymore, its bias is also effectively dead. As batch normalization can contain a considerable amount of parameters and hence expressive power \citep{frankle_training_2020}, we made all normalization affine. For the residual networks, we did not prune residual connections, as they do not contain a weight.

\section{Distance functions}\label{sec:distance_functions}

\begin{figure}
\centering
\includegraphics[width=\textwidth]{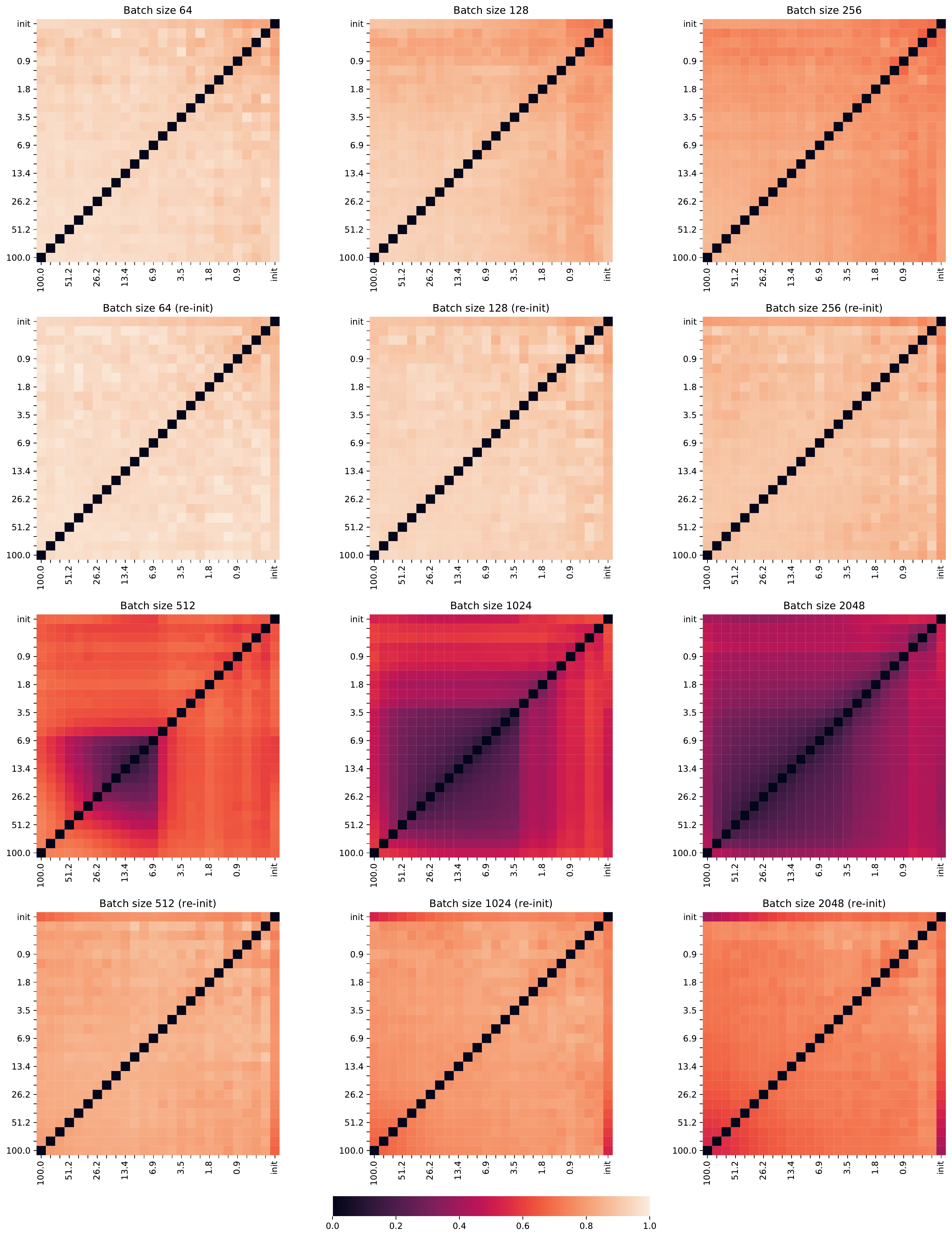}
\caption{Angular distances between the retrained networks troughout sparsity. Last row/column compares with the network at initialization. Experiments on CIFAR-10 with ResNet20.}\label{figure:heatmaps_full}
\end{figure}

\begin{figure}
\centering
\includegraphics[width=\textwidth]{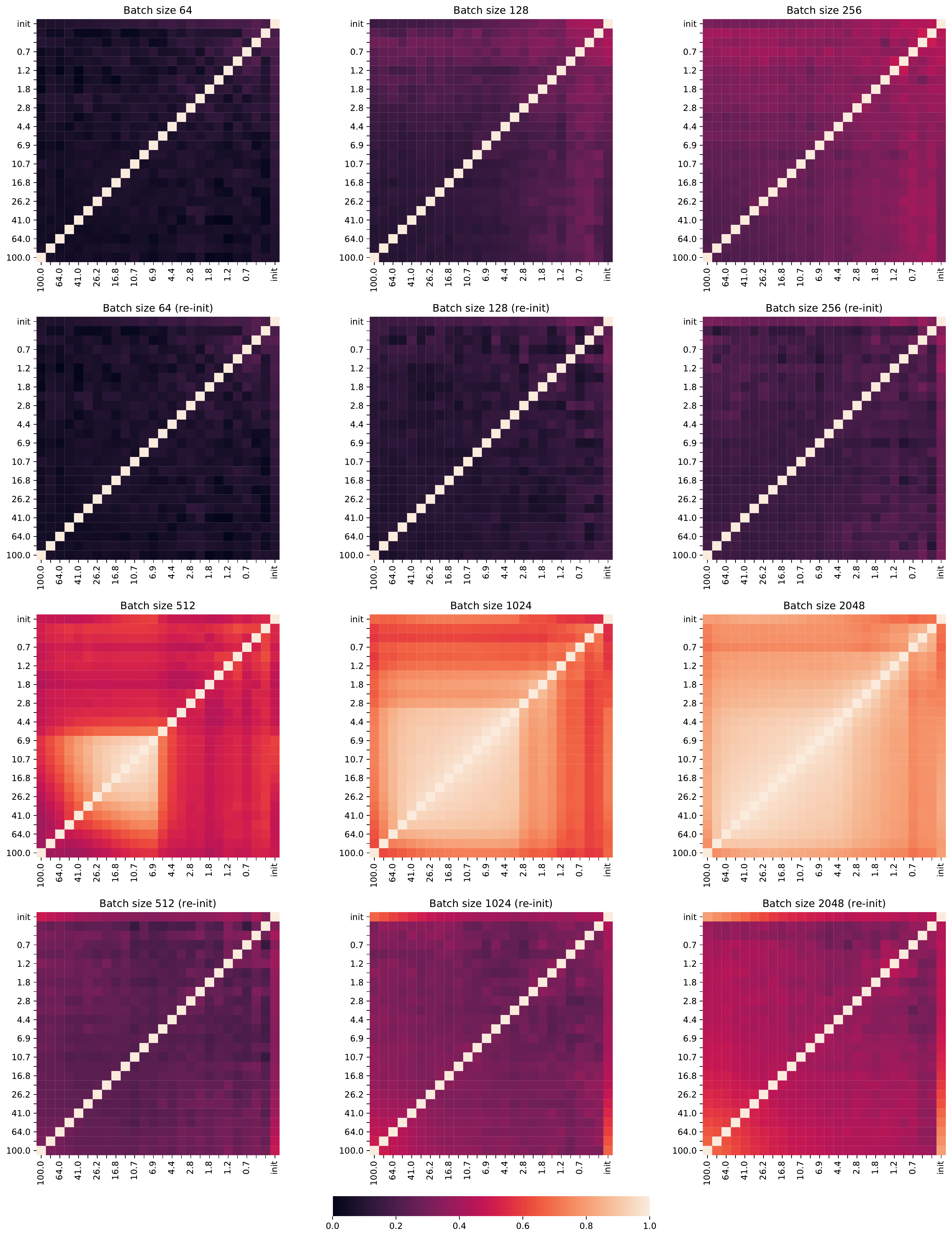}
\caption{Cosine distances between the retrained networks troughout sparsity. Last row/column compares with the network at initialization. Experiments on CIFAR-10 with ResNet20.}\label{figure:heatmaps_cos}
\end{figure}

\begin{figure}
\centering
\includegraphics[width=\textwidth]{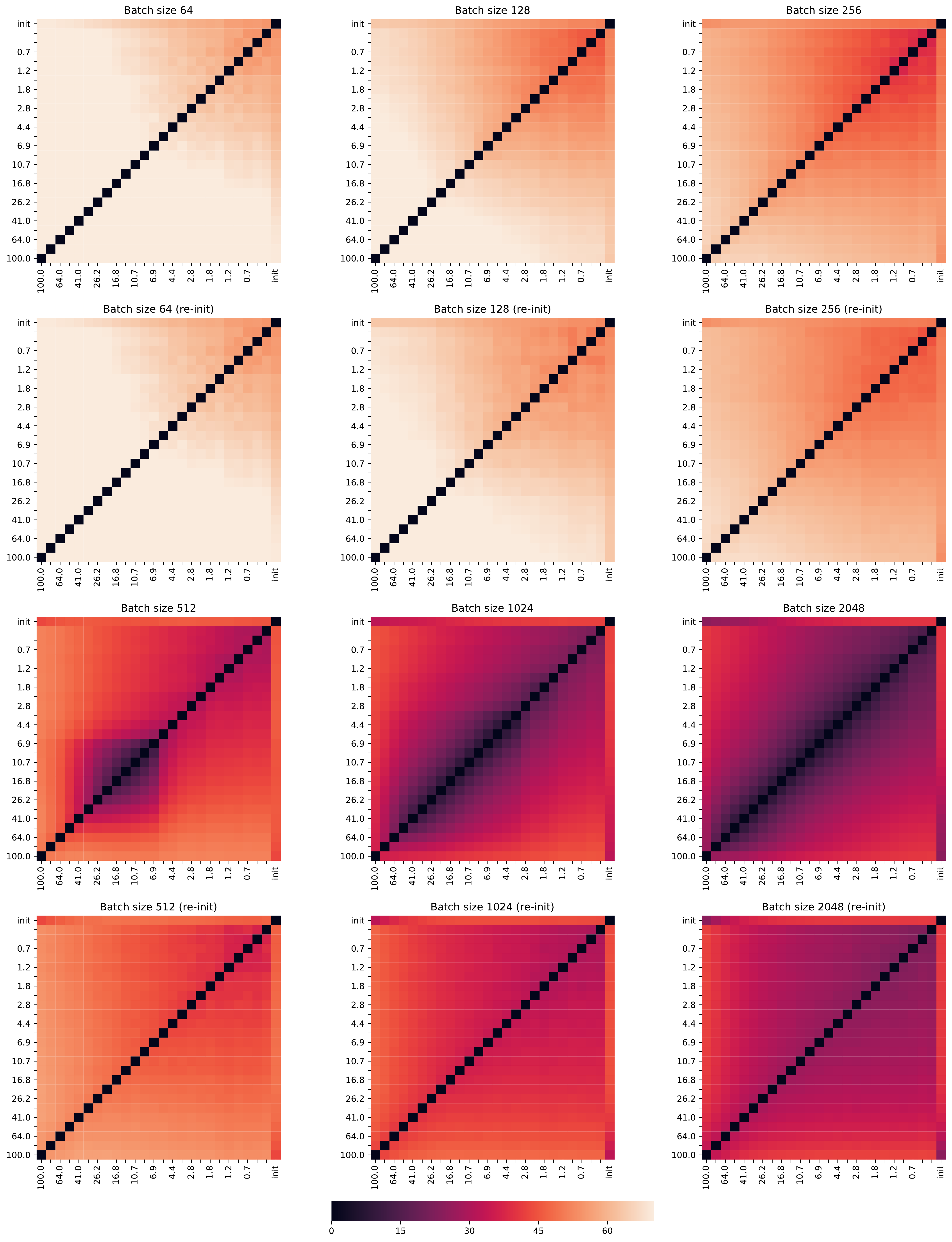}
\caption{Euclidean distances between the retrained networks troughout sparsity. Last row/column compares with the network at initialization. Experiments on CIFAR-10 with ResNet20.}\label{figure:heatmaps_l2}
\end{figure}

In section~\ref{sec:similarity} we chose the angle to quantify the distance between networks. Here we highlight two alternatives: the cosine distance in figure~\ref{figure:heatmaps_cos} and Euclidean distance in figure~\ref{figure:heatmaps_l2}. Figure~\ref{figure:heatmaps_full} displays the full set of similarity matrices for the angular distances.

The same overall pattern manifests in all 3 figures. At low batch sizes, the networks are very dissimilar to each other. While at high sparsities the networks become increasingly similar. 

\section{Extra material on alternative stabilization}\label{app:alternative_stabilization}

To support the generality of our results, we conducted the same stabilization experiments on various architectures. 

\begin{figure}
\centering
\includegraphics[width=\textwidth]{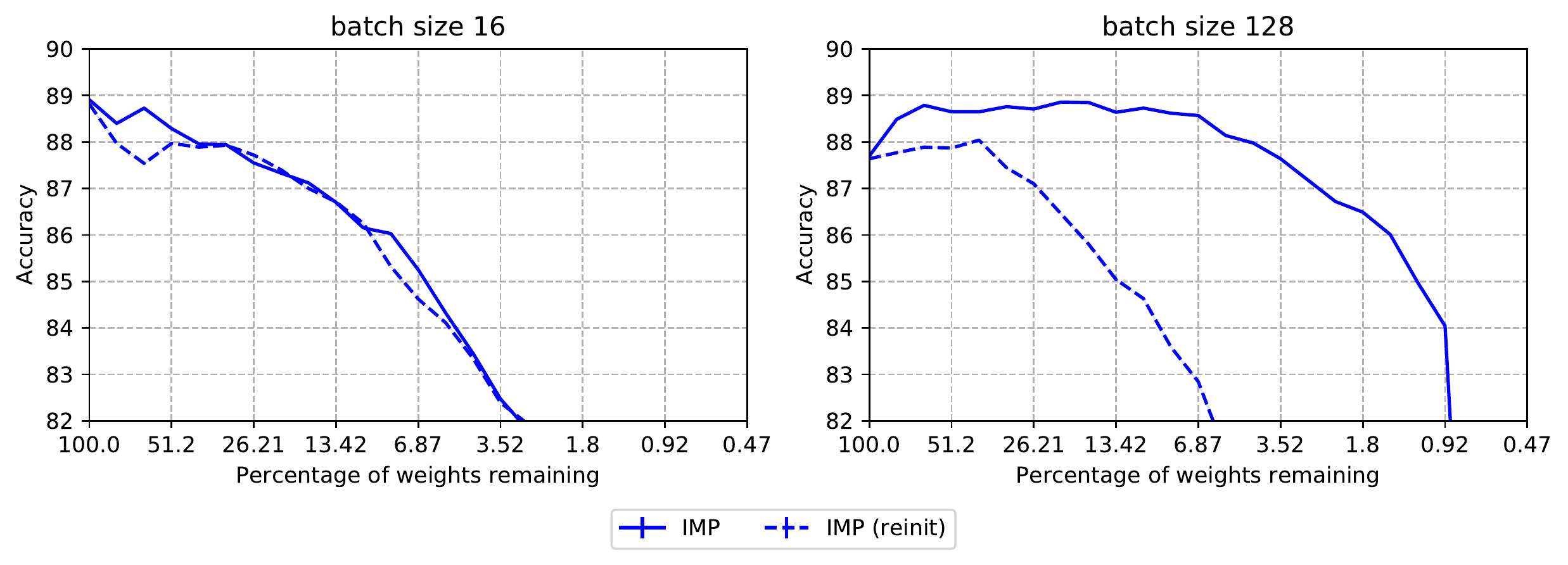}
\caption{Influence of the batch size on IMP. Experiments on CIFAR-10 with Conv-4.}\label{figure:big_batch_conv4}
\end{figure}

First, we look at the Conv-4 architecture. \cite{frankle_lottery_2019} show that this architecture is already stable at initialization with various optimizers and hyperparameters. We first show that we can destabilize learning, by picking a low batch size and fairly high weight decay (0.01). To further encourage instability, we also use a stepwise learning rate schedule (10x drops at epoch 55 and 65) instead of a cosine schedule; this means the network trains much longer at the highest learning rate. This prevents IMP from finding lottery tickets; see figure~\ref{figure:big_batch_conv4} on the left. Next, we increase the learning rate back, which re-stabilizes IMP; see figure~\ref{figure:big_batch_conv4} on the right.

Next, we consider more large-scale settings, like ResNet50 on ImageNet. We calculated that a minimal reproduction of figure~\ref{figure:big_batch_sweep} would take over 350 days on our hardware\footnote{This assumes 4 IMP runs of 25 rounds on 2 GPUs with about 7 days to train a single model on ImageNet on 1 GPU.}, which was deemed infeasible. Hence we provide a more limited indicative experiment. We copy the one-shot IMP performance from \cite{frankle_linear_2020} at 30\% sparsity. Next, we perform the same experiment at a lower learning rate\footnote{We could have equivalently increased the batch size, however we did not possess sufficient memory for this. So instead of applying gradient accumulation, we simply lowered the learning rate.}. 

\begin{table}[h!]
\centering
\begin{tabular}{l l l}
\toprule
Learning rate & Dense accuracy & Accuracy at 30\% sparsity \\\hline
 0.4 & 76.1\% & -2.4\% \\
 0.1  & 72.1\% & -2.0\% \\
 \bottomrule
\end{tabular}\label{table:resnet50}
\end{table}

The small decrease in learning rate is not sufficient to fully stabilize IMP, but we note that the accuracy gap does decrease. We had no resources to experiment with even lower learning rates. For the sake of full clarity, we mention that there are some minor differences between our experiment, and the version of \cite{frankle_linear_2020}. First, we use the face-blurred ILSVRC ImageNet, while \cite{frankle_linear_2020} do not disclose what exact version. Second, we apply affine batch normalization instead of the regular version.

\section{Extra material on forced repellence}\label{app:forced_repellence}

\begin{figure}
\centering
\includegraphics[width=0.6\textwidth]{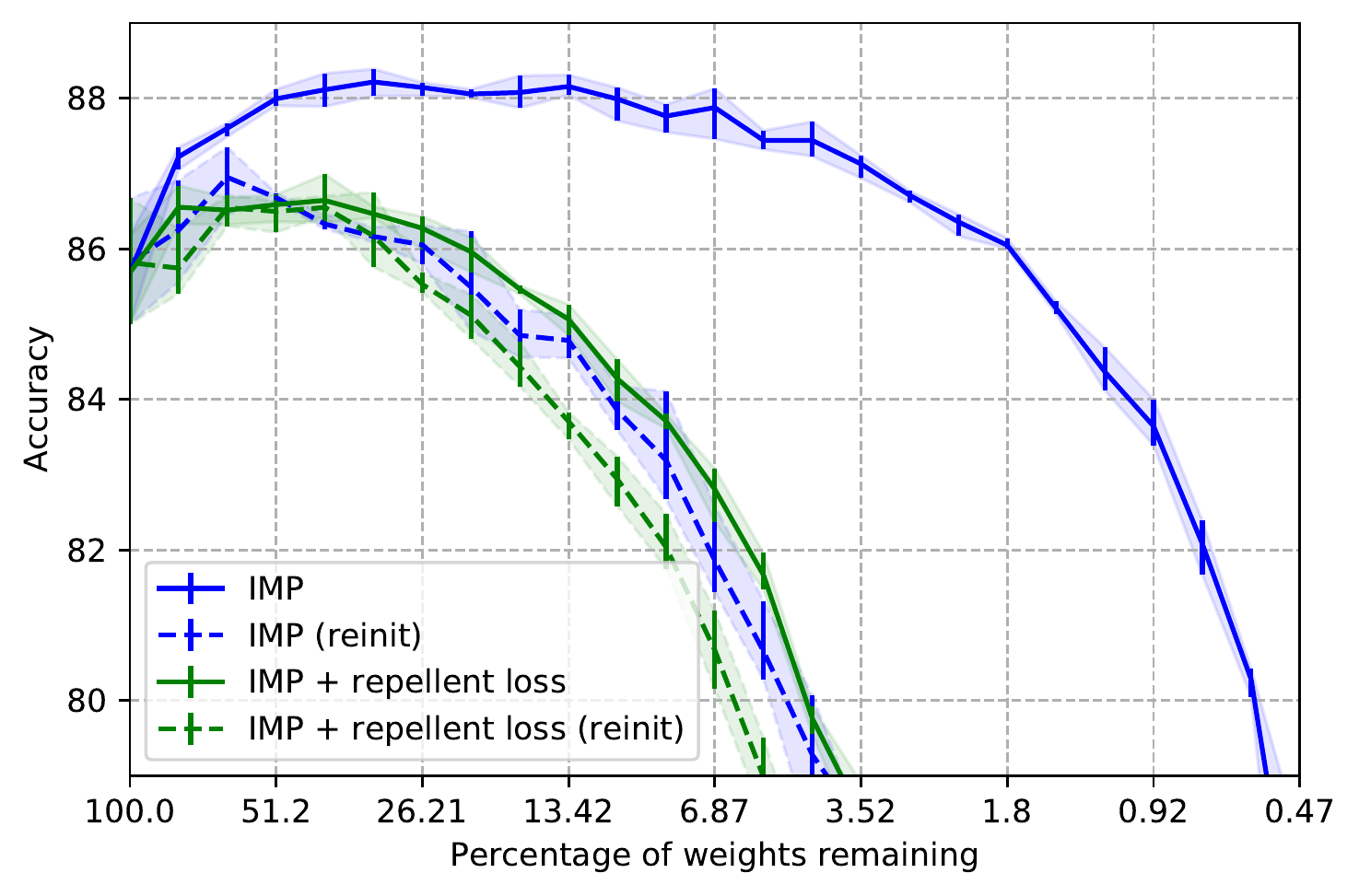}
\caption{Comparison of repellent IMP with regular IMP. Repellent loss ($\lambda_r=2$). Experiments on CIFAR-10 with Conv-4. Average of 3 runs with different seeds; error bars indicate maximum and minimum values.}\label{figure:IMP_repellent_conv4}
\end{figure}

\begin{figure}
\centering
\includegraphics[width=0.6\textwidth]{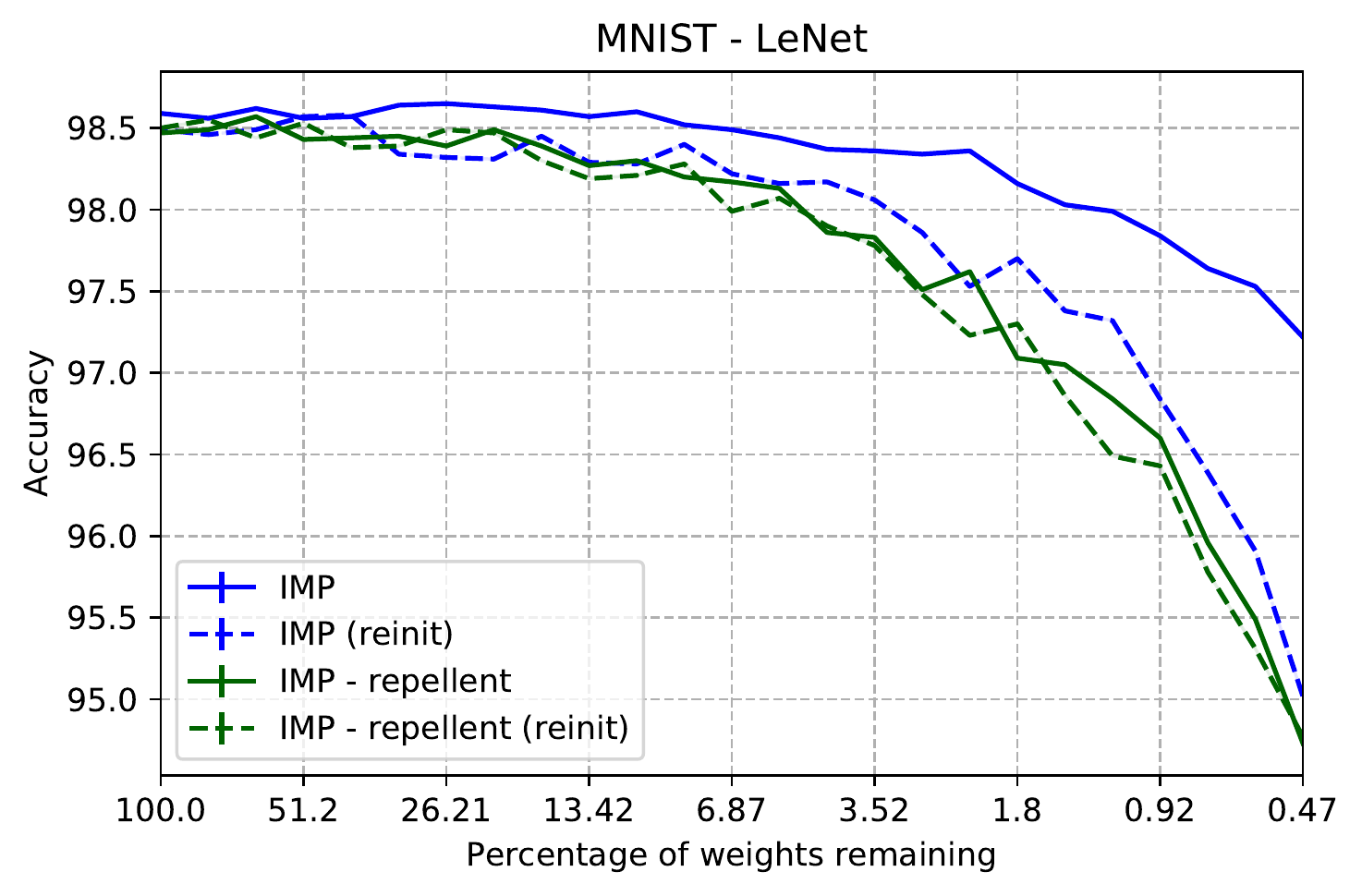}
\caption{Comparison of repellent IMP with regular IMP. Repellent loss ($\lambda_r=2$). Experiments on MNIST with LeNet.}\label{figure:IMP_repellent_lenet}
\end{figure}

\begin{figure}
\centering
\includegraphics[width=0.8\textwidth]{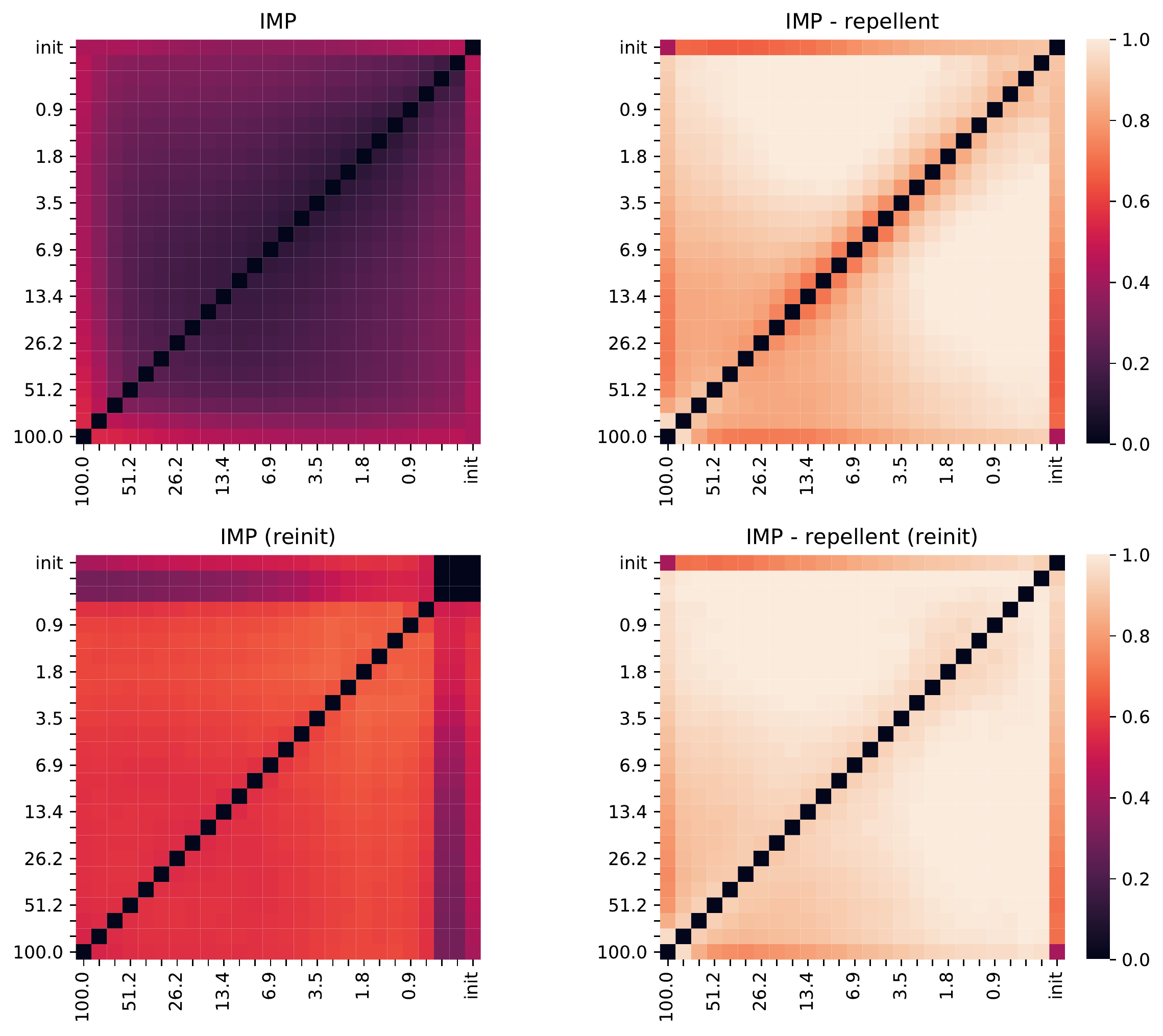}
\caption{Angular distances for the repellent runs between the retrained networks troughout sparsity. Last row/column compares with the network at initialization. Experiments on CIFAR-10 with Conv-4.}\label{figure:heatmaps_repellent_conv4}
\end{figure}

We repeat the experiment on forced repellence of figure~\ref{figure:IMP_repellent} on the Conv-4 architecture in figure~\ref{figure:IMP_repellent_conv4} with very similar results. There is a very slight improvement of the repellent IMP (in green) over the regular IMP reinitialization (in dashed blue). This can be explained by the fact that $\lambda_r$ is slightly too weak (we did not re-tune it). We confirm this by looking at the distances in figure~\ref{figure:heatmaps_repellent_conv4}. In the top right, the distances are still noticeable higher near the diagonal than usual (bottom right). In figure~\ref{figure:IMP_repellent_lenet}, we also apply forced repellence on LeNet.

\begin{figure}
\centering
\includegraphics[width=0.8\textwidth]{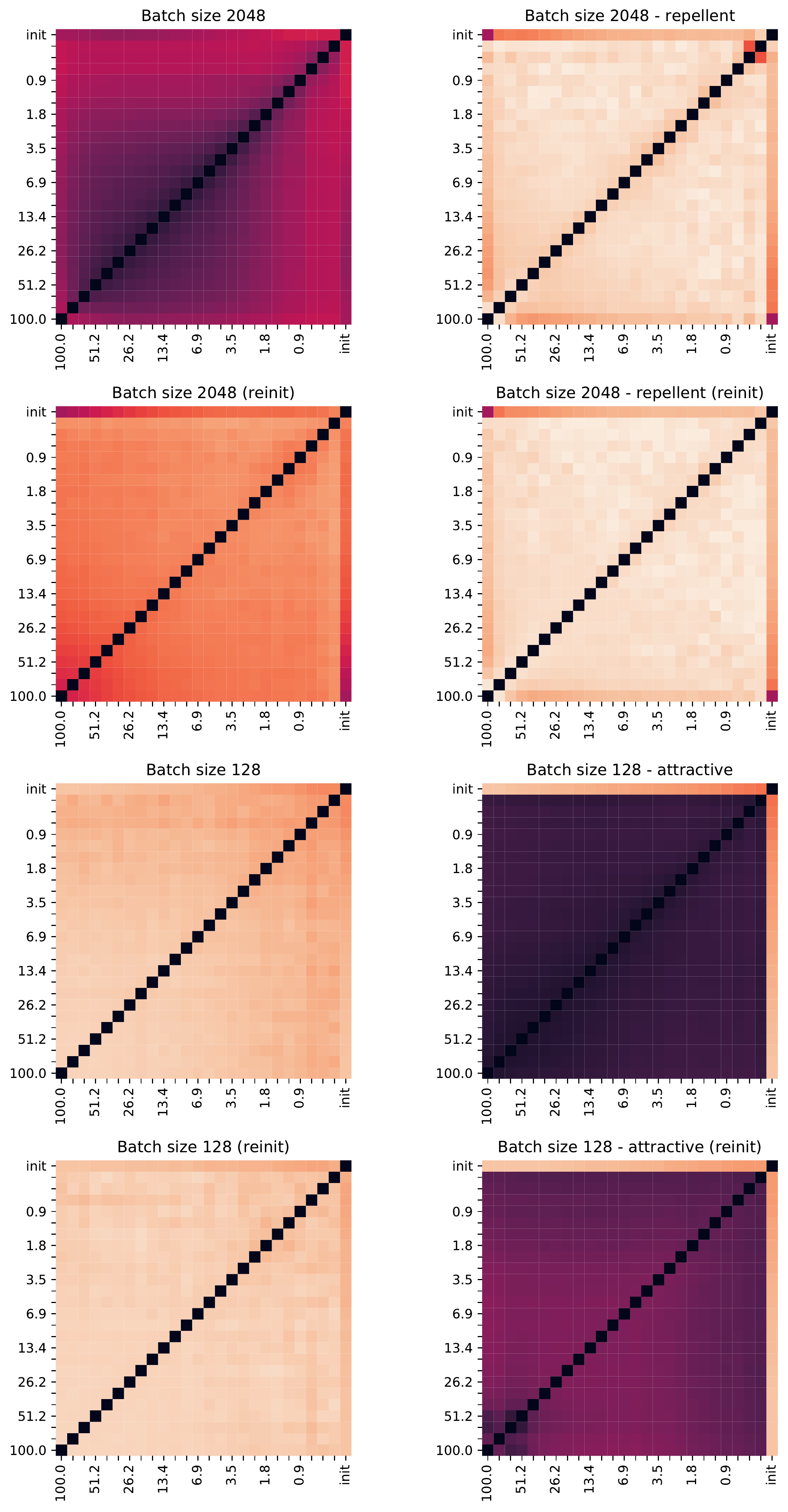}
\caption{Angular distances for the repellent runs  between the retrained networks troughout sparsity. Last row/column compares with the network at initialization. Experiments on CIFAR-10 with ResNet20.}\label{figure:heatmaps_repellent}
\end{figure}

\begin{figure}
\centering
\includegraphics[width=0.5\textwidth]{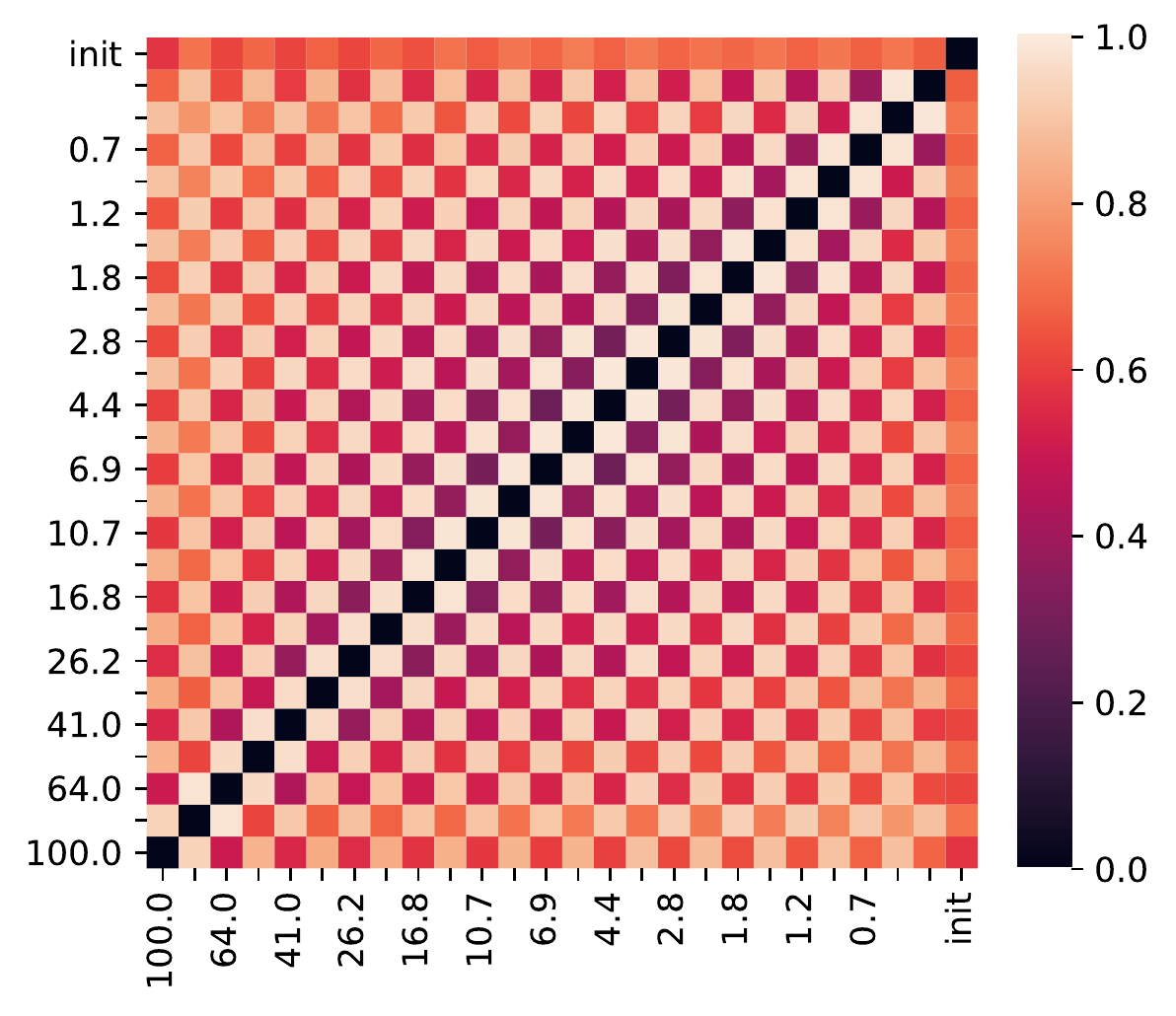}
\caption{Using a simplified repellent loss that only considers the previous solution results in biphasal retraining. Experiment on MNIST with LeNet.}\label{figure:heatmap_checkerboard}
\end{figure}

Next, we also visualized the distance matrices for the repellent runs; see~\ref{figure:heatmaps_repellent}. Lastly, we show that a constant memory repellent loss that only repels from the previous solution is not sufficient. We show the distances in figure~\ref{figure:heatmap_checkerboard}. Clearly, the model retrains the one but last model. This is also reflected in the accuracy: these runs can still find lottery tickets, but performance is less good than in the normal case.

\section{Extra material on quantifying similarity}\label{app:quantifying similarity}

\begin{figure}
\centering
\includegraphics[width=\textwidth]{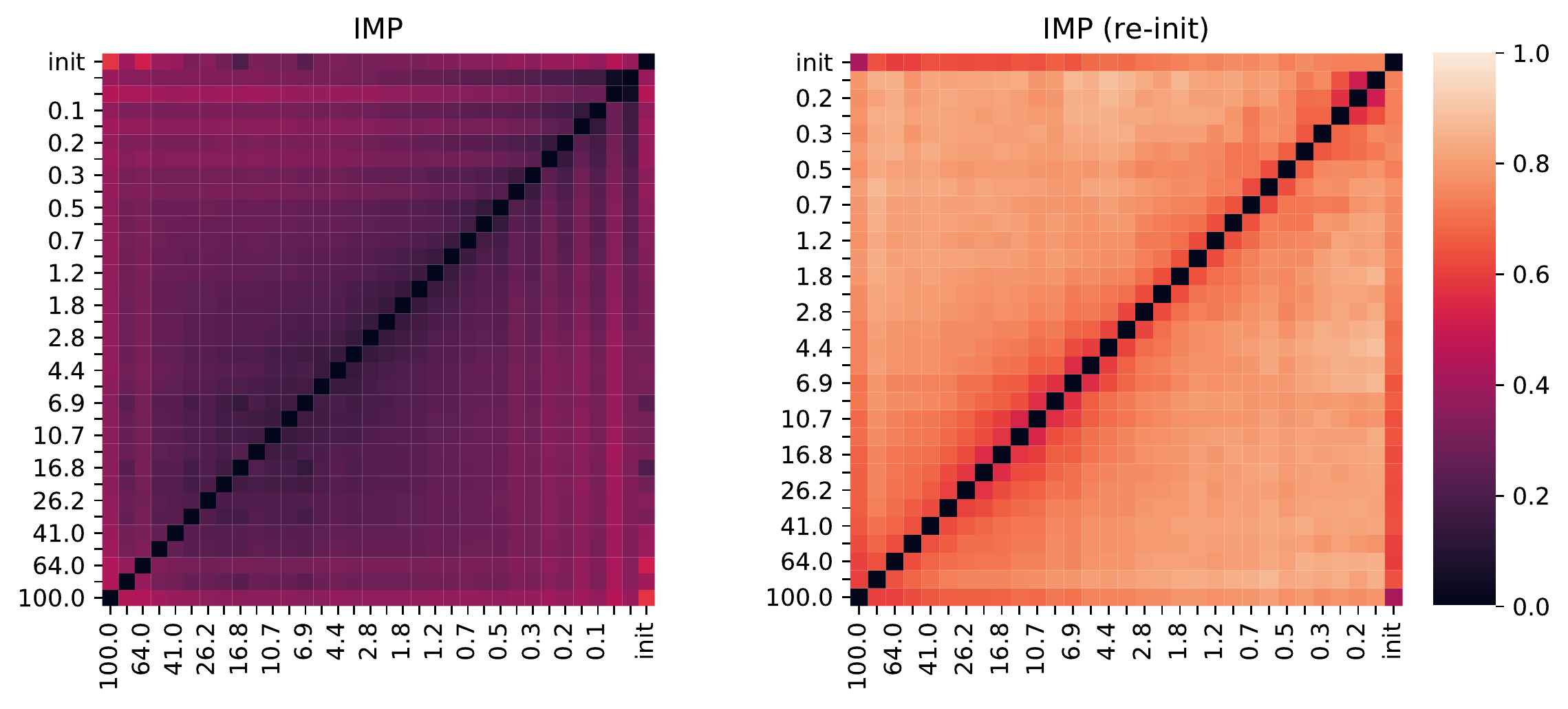}
\caption{Angular distances between the retrained networks troughout sparsity. Last row/column compares with the network at initialization. Experiments on MNIST with LeNet.}\label{figure:heatmaps_mnist}
\end{figure}

\begin{figure}
\centering
\includegraphics[width=0.8\textwidth]{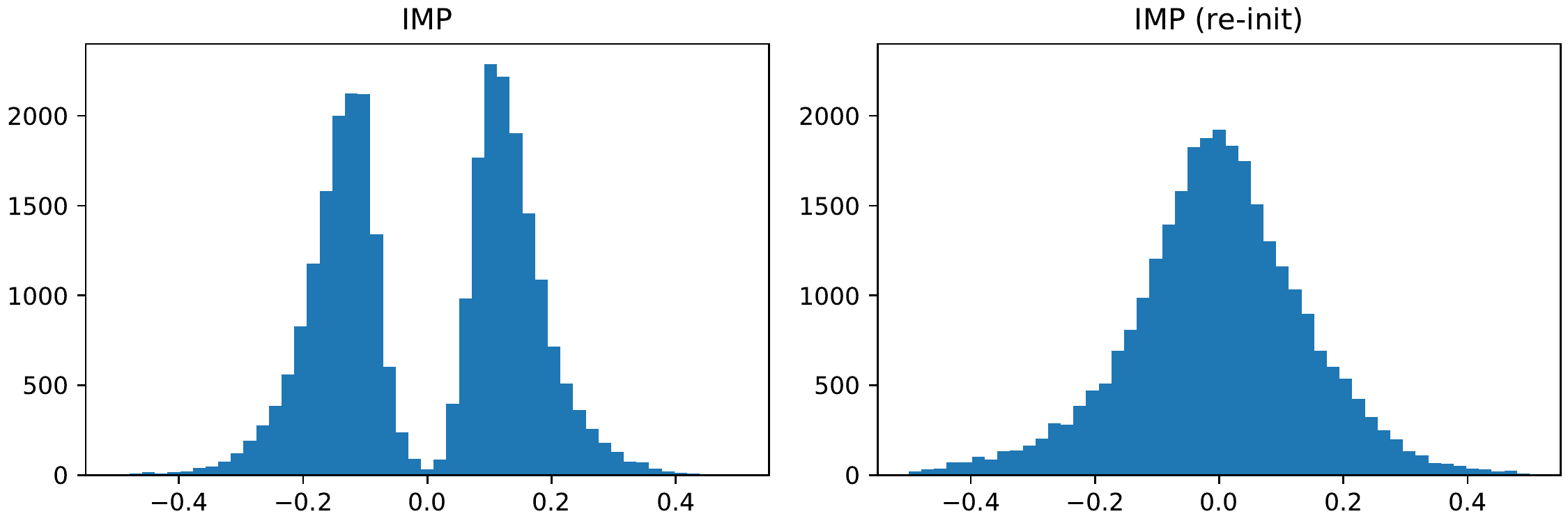}
\caption{Histograms of the model weights after 10 pruning iterations (at sparsity 10.7\%). Experiments on MNIST with LeNet.}\label{figure:histograms_mnist}
\end{figure}

Here, we also include the distance matrices for our experiments on other architectures. Figure~\ref{figure:heatmaps_mnist} displays the distances using IMP on MNIST with LeNet. The corresponding weight histograms can be found in figure~\ref{figure:histograms_mnist}.

\begin{figure}
\centering
\includegraphics[width=0.8\textwidth]{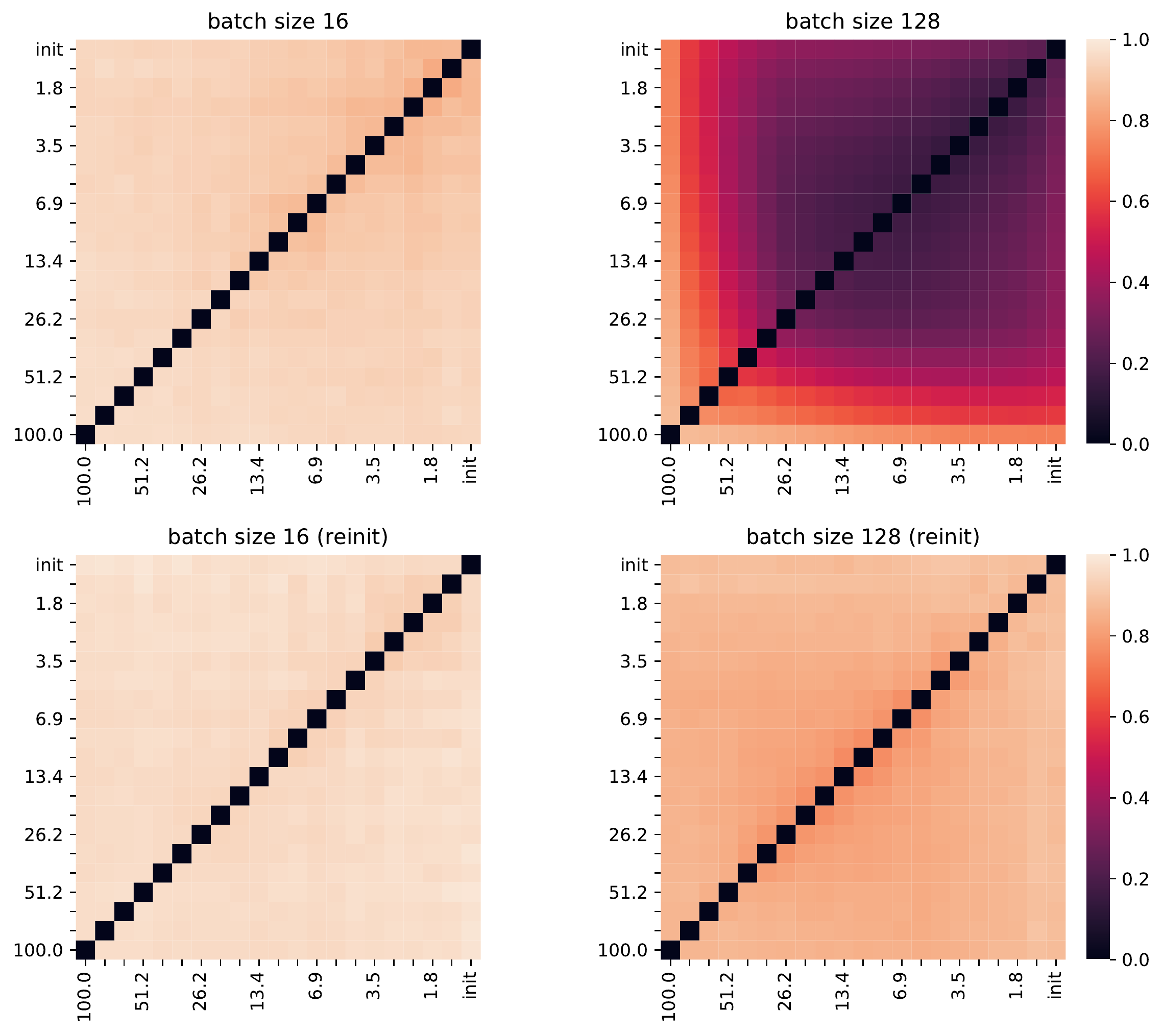}
\caption{Angular distances between the retrained networks troughout sparsity. Last row/column compares with the network at initialization. Experiments on CIFAR-10 with Conv-4.}\label{figure:heatmaps_conv4}
\end{figure}

\begin{figure}
\centering
\includegraphics[width=\textwidth]{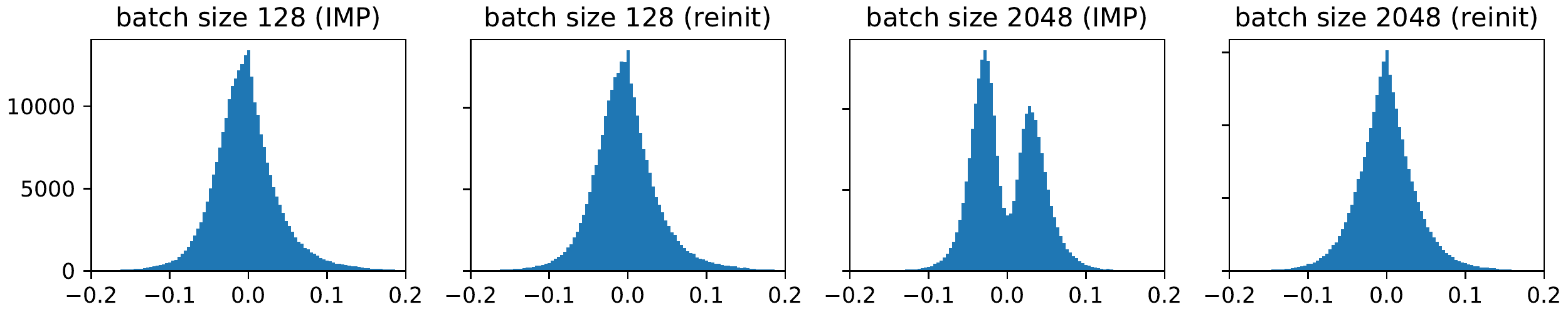}
\caption{Histograms of the model weights after 10 pruning iterations (at sparsity 10.7\%). Experiments on CIFAR-10 with Conv-4.}\label{figure:histograms_conv4}
\end{figure}

\begin{figure}
\centering
\includegraphics[width=0.6\textwidth]{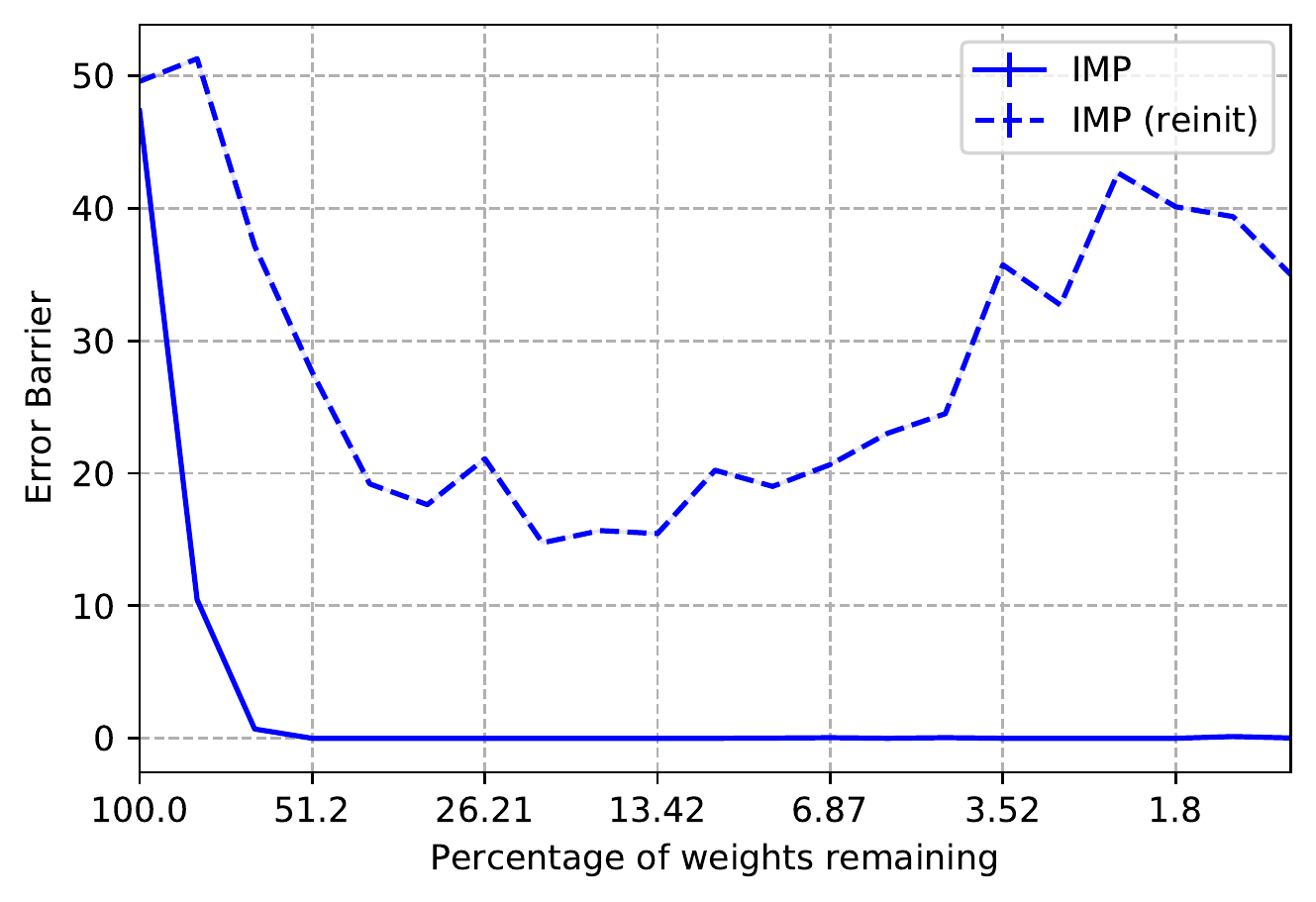}
\caption{The maximum error increase on the linear path over sparsities. Experiments on CIFAR-10 with Conv-4, with a batch size of 128 and no warm-up.}\label{figure:error_barrier_conv4}
\end{figure}

In figure~\ref{figure:heatmaps_conv4}, we display the distances from the stabilization experiments on Conv-4 from figure~\ref{figure:big_batch_conv4}. The 4 last sparsities are left out because the network failed to train for batch size 16. Figure~\ref{figure:histograms_conv4} contains the histograms. Due to the high weight decay, we can see that the weights are mostly contained on a very narrow band and the distribution is more peaked than usual. Still, the stable IMP run (middle right) maintains its bimodal distribution. Curiously, the modes have differing heights in this case. Lastly, figure~\ref{figure:error_barrier_conv4} shows the error barrier for the regular IMP runs on Conv-4 (so not with high weight decay).

\begin{figure}
\centering
\includegraphics[width=0.6\textwidth]{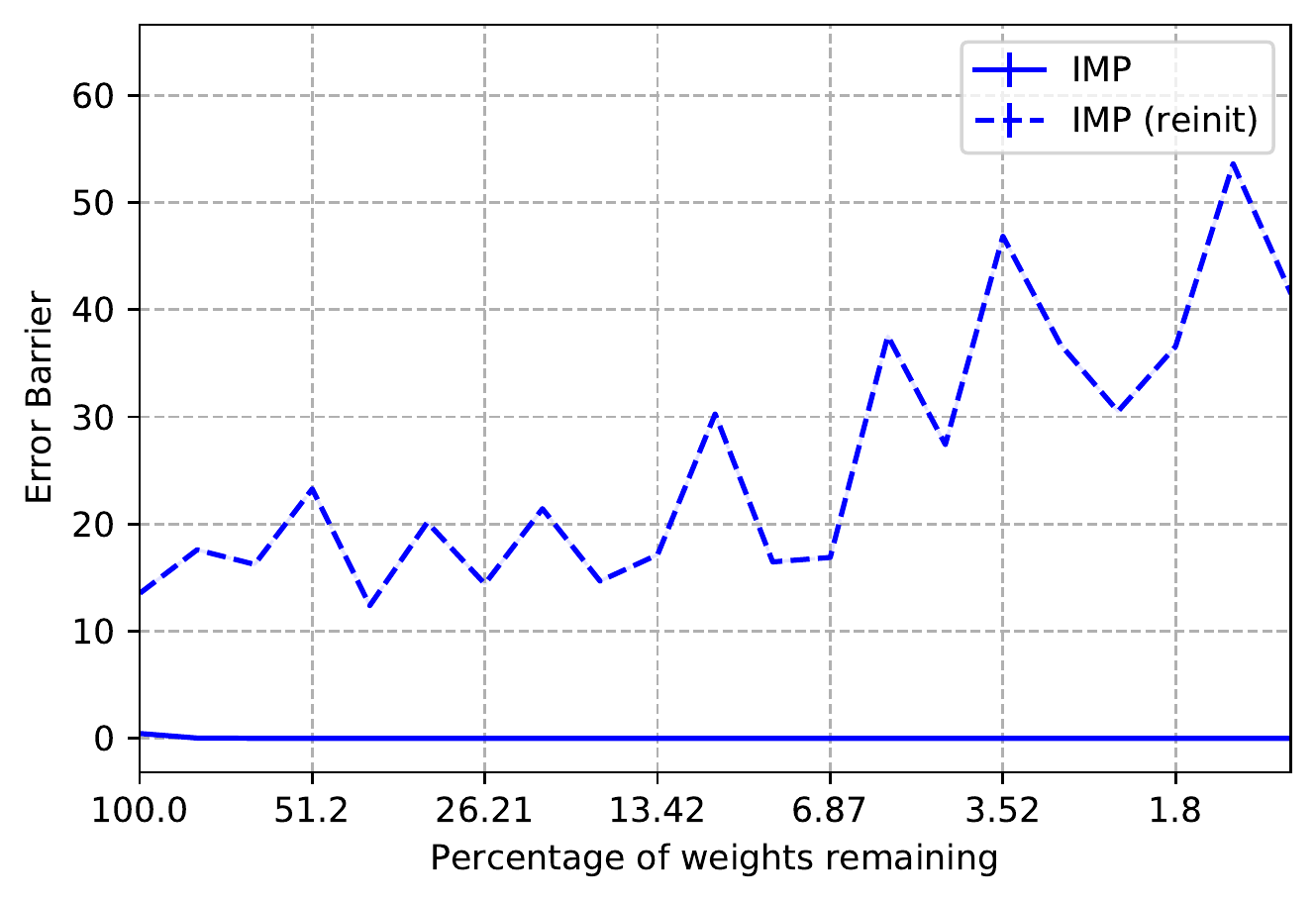}
\caption{The maximum error increase on the linear path over sparsities. Experiments on MNIST with LeNet.}\label{figure:error_barrier_lenet}
\end{figure}

\end{document}